\documentclass[journal]{IEEEtran}
\usepackage{xcolor}
\usepackage{amsmath,amssymb,amsfonts}
\usepackage{algorithmic}
\usepackage{graphicx}
\usepackage{textcomp}
\usepackage{threeparttable}
\usepackage{array}
\usepackage{siunitx}
\usepackage{booktabs}
\usepackage[switch]{lineno} 
\usepackage{xurl}
\usepackage{multirow}
\usepackage[caption=false,font=normalsize,labelfont=sf,textfont=sf]{subfig}
\usepackage{url}
\usepackage{verbatim}
\usepackage{balance}

\begin{document}

\title{Cross-Sequence Semi-Supervised Learning for Multi-Parametric MRI-Based Visual Pathway Delineation}

\author{Alou Diakite, Cheng Li, Lei Xie, Yuanjing Feng, Ruoyou Wu, Jianzhong He, Hairong Zheng, and Shanshan Wang
\thanks{This research was partly supported by the Shenzhen Medical Research Fund (B2402047), National Natural Science Foundation of China (62222118, U22A2040), Shenzhen Science and Technology Program (JCYJ20220531100213029), and Key Laboratory for Magnetic Resonance and Multimodality Imaging of Guangdong Province (2023B1212060052). (Corresponding author: Shanshan Wang, e-mail: ss.wang@siat.ac.cn) }
\thanks{Alou Diakite and Ruoyou Wu are with Paul C. Lauterbur Research Center for Biomedical Imaging, Shenzhen Institutes of Advanced Technology, Chinese Academy of Sciences, Shenzhen, 518055, China, and are also with University of Chinese Academy of Sciences, Beijing, 100040, China (e-mail: aloudiakite@siat.ac.cn; ry.wu@siat.ac.cn).}
\thanks{Cheng Li, Hairong Zheng, and Shanshan Wang are with Paul C. Lauterbur Research Center for Biomedical Imaging, Shenzhen Institutes of Advanced Technology, Chinese Academy of Sciences, Shenzhen, 518055, China (e-mail: cheng.li6@siat.ac.cn; hr.zheng@siat.ac.cn; e-mail: ss.wang@siat.ac.cn).}
\thanks{Lei Xie, Yuanjing Feng, and Jianzhong He are with Zhejiang University of Technology, Hangzhou, 310023, China (e-mail: leix@zjut.edu.cn; fyjing@zjut.edu.cn; hjzhong@zjut.edu.cn).}
}

\markboth{Journal of \LaTeX\ Class Files,~Vol.~14, No.~8, August~2025}%
{Shell \MakeLowercase{\textit{et al.}}: A Sample Article Using IEEEtran.cls for IEEE Journals}


\maketitle
\begin{abstract}
Accurately delineating the visual pathway (VP) is crucial for understanding the human visual system and diagnosing related disorders. Exploring multi-parametric MR imaging data has been identified as an important way to delineate VP. However, due to the complex cross-sequence relationships, existing methods cannot effectively model the complementary information from different MRI sequences. In addition, these existing methods heavily rely on large training data with labels, which is labor-intensive and time-consuming to obtain. In this work, we propose a novel semi-supervised multi-parametric feature decomposition framework for VP delineation. Specifically, a correlation-constrained feature decomposition (CFD) is designed to handle the complex cross-sequence relationships by capturing the unique characteristics of each MRI sequence and easing the multi-parametric information fusion process. Furthermore, a consistency-based sample enhancement (CSE) module is developed to address the limited labeled data issue, by generating and promoting meaningful edge information from unlabeled data. We validate our framework using two public datasets, and one in-house Multi-Shell Diffusion MRI (MDM) dataset. Experimental results demonstrate the superiority of our approach in terms of delineation performance when compared to seven state-of-the-art approaches.
\end{abstract}

\begin{IEEEkeywords}
Visual pathway, feature decomposition, semi-supervised learning, multi-parametric MR imaging.
\end{IEEEkeywords}

\section{Introduction}
\label{sec:introduction}
\IEEEPARstart{T}{he} visual pathway (VP) consists of structures such as the optic nerve, optic chiasm, optic tract, and others, which transmit visual information from the retina to the brain \cite{Mansoor2016}. Therefore, accurate delineation of the visual pathway (VP) is crucial for understanding the human visual system \cite{takemura2019diffusivity} and can lead to better clinical outcomes in visual disorders \cite{ma2016preoperative}. However,  VP structures can exhibit significant variability in size, shape, and orientation among individuals \cite{Li2021}. This variability and complex connectivity make the manual delineation of these structures more challenging and time-consuming, so automated methods are desired \cite{liang2020multi, huang20213d}. Two categories of techniques have emerged for automated VP delineation: classical methods and deep learning (DL) based methods. These techniques provide powerful tools for accurately delineating VPs, benefiting researchers and clinicians in clinical practice.

Classical methods, including both atlas-based \cite{Harrigan2014} and model-based \cite{Bekes2008} approaches, are used for VP delineation. Atlas-based methods involve computing a transformation between a reference image volume (atlas) and a target image using image registration. This transformation is then used to project labels from the atlas onto the image volume for delineation \cite{Noble2011, Yang2014}. Multi-atlas-based methods use a database of atlas images and fuse the projected labels to produce the final delineation result. However, these conventional methods mostly rely on large sample data and shallow feature extraction, which may limit the delineation performance. 

DL techniques have emerged as a powerful tool for medical image delineation, including VP delineation \cite{ huang20213d}. These DL-based methods have undergone three periods. At first, deep learning methods were solely employed for feature extraction, while delineation relied on conventional techniques. Mansoor et al.\cite{Mansoor2016}, for example, utilized stacked autoencoders to learn new feature representations for a model-based delineation approach. Deep learning was also applied to pixel/voxel classification in the subsequent period. For instance, Ren et al. \cite{Ren2018} extended the original CNN to an interleaved structure for joint delineation of the optic nerve and chiasm. Liang et al. \cite{liang2020multi} introduced a multi-view spatial aggregation method for joint localization and delineation of optic nerve and optic chiasm. Moreover, U-Net network-based approaches \cite{Ronneberger2015} that integrate feature extraction and pixel/voxel classification procedures have emerged for VP delineation from CT/MRI images without hand-crafted features \cite{Ai2020}. Furthermore, Chen et al. \cite{chen2023adaptive} proposed an adaptive region-specific loss scheme for improved optic nerve and optic chiasm delineation. Though these methods have shown promising results, they are also limited to delineating only one or two structures associated with VP (optic nerve, optic chiasm) and typically use single-sequence data, which may not capture the complete information needed for accurate delineation.

\begin{figure*}[h]
  \centering
  \subfloat[]{\includegraphics[width=0.2\linewidth]{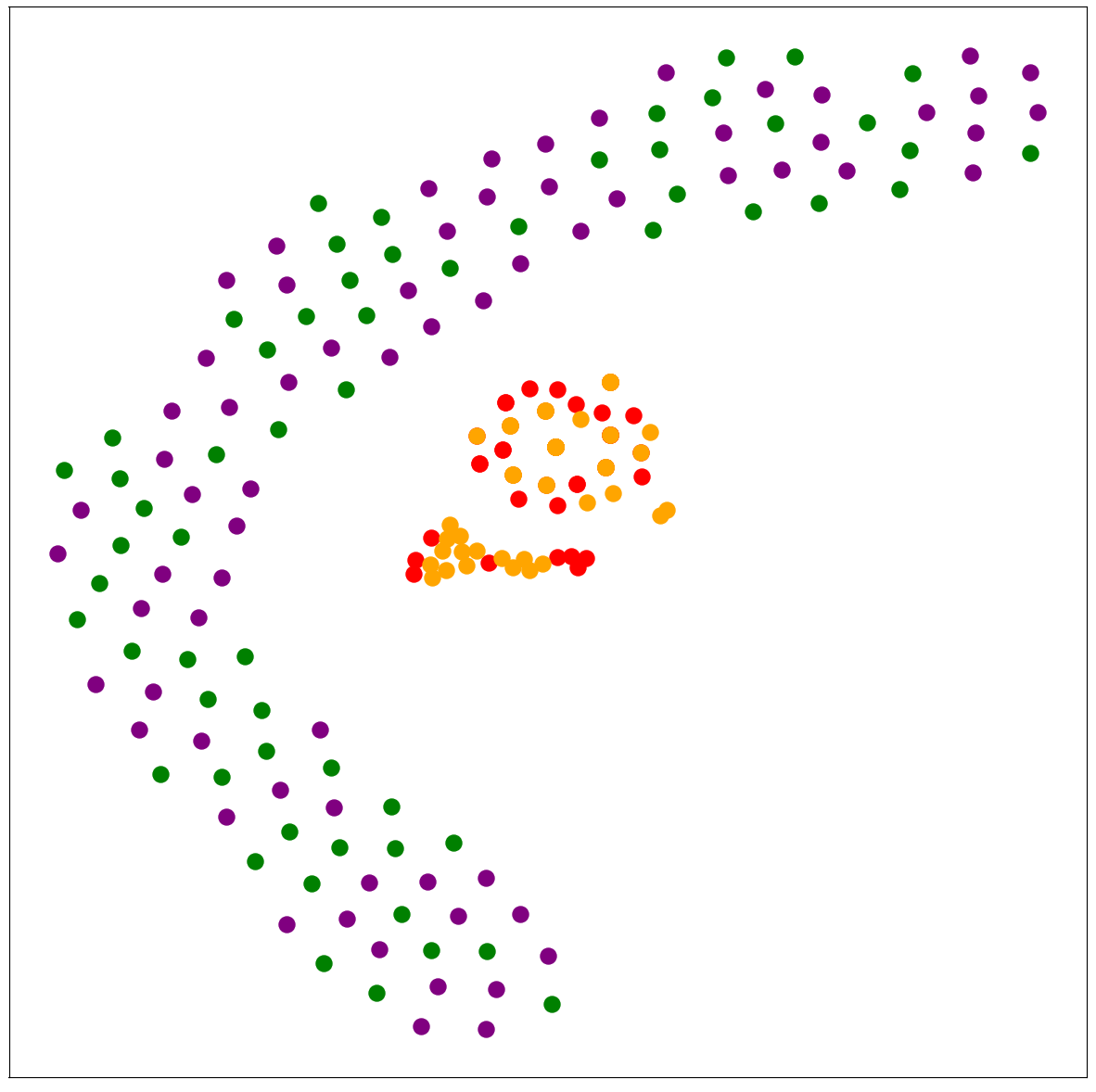}\label{fig1:a}}
  \hfill
  \subfloat[]{\includegraphics[width=0.2\linewidth]{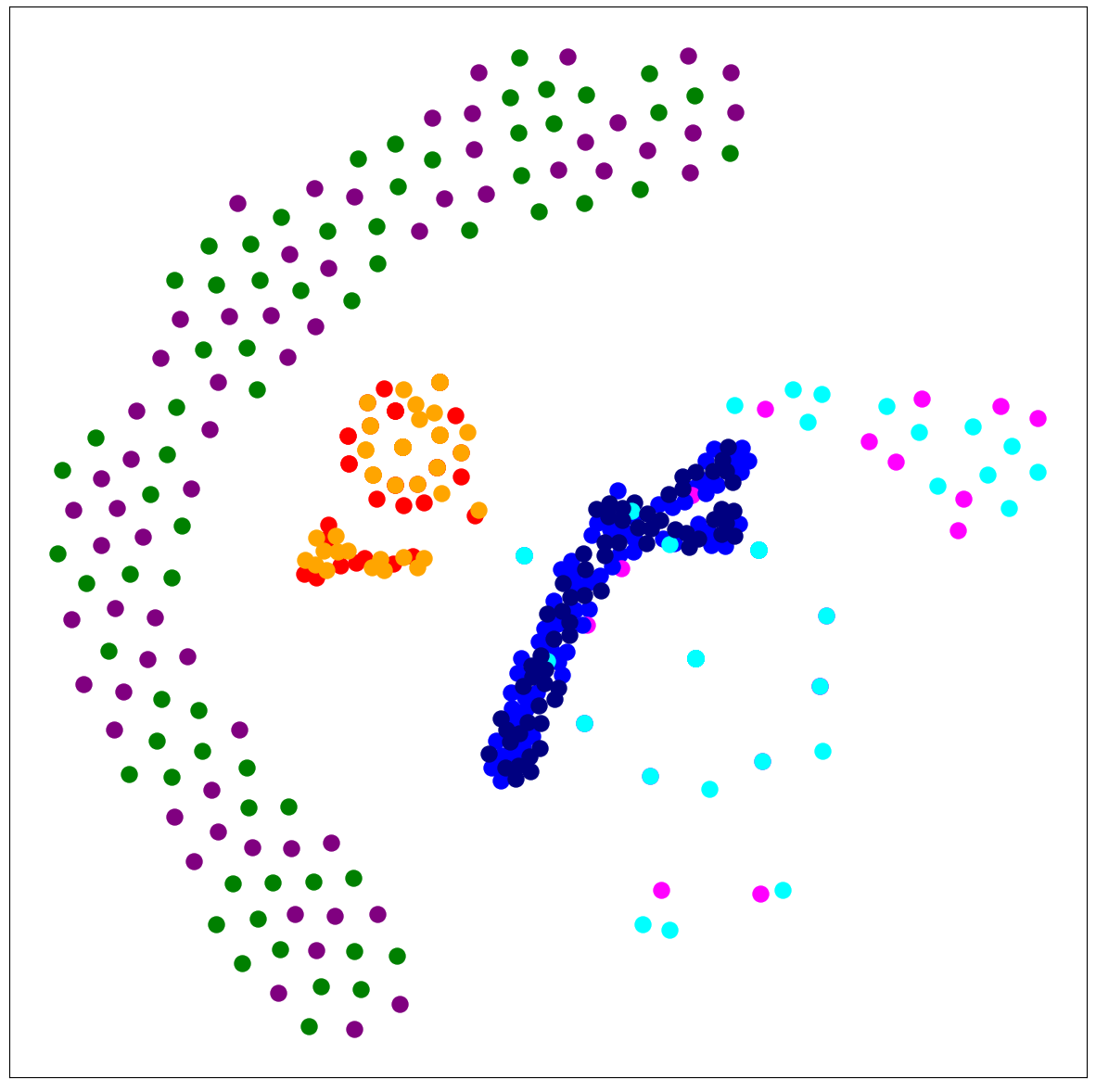}\label{fig1:b}}
  \hfill
  \subfloat[]{\includegraphics[width=0.2\linewidth]{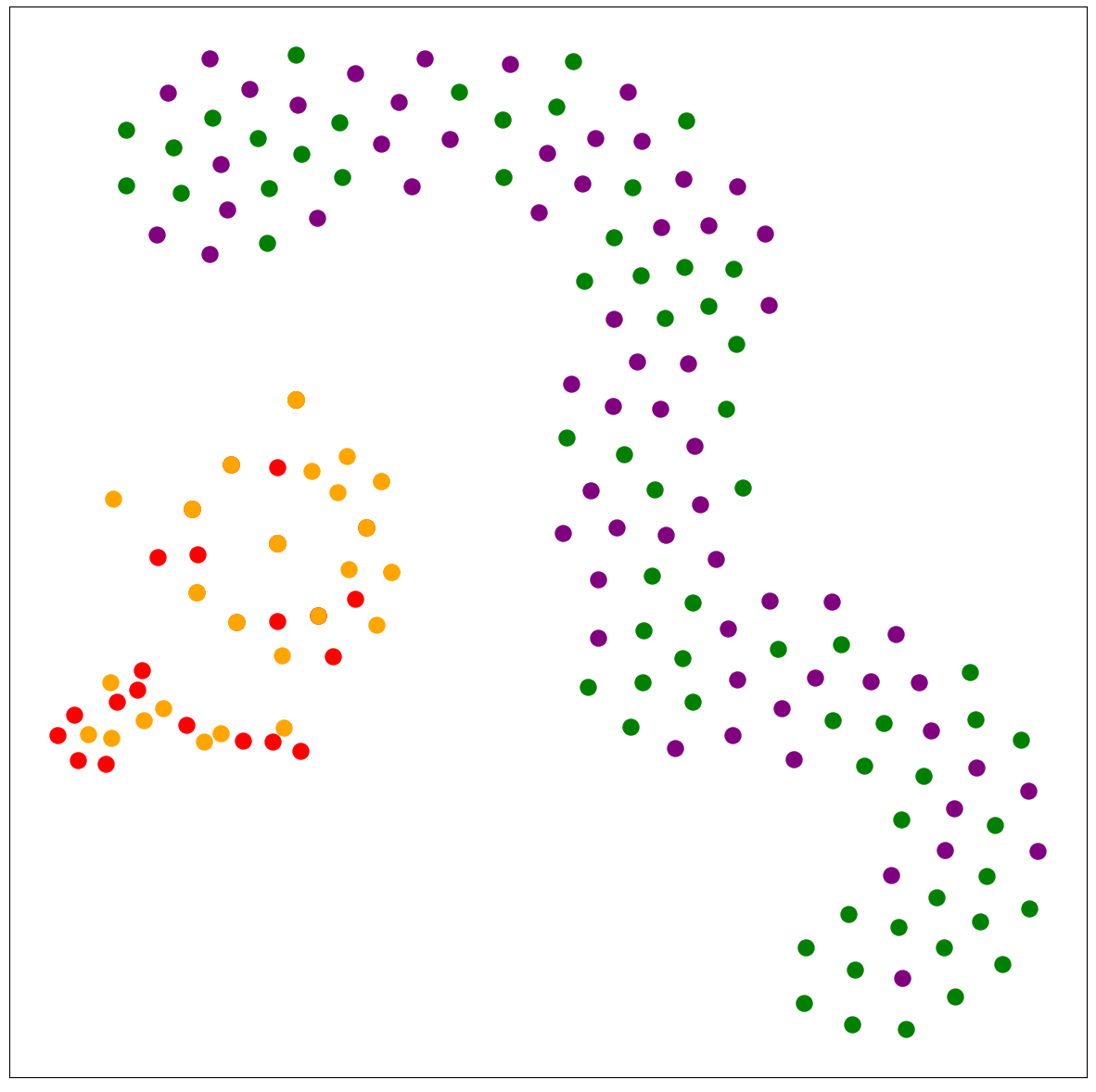}\label{fig1:c}}
  \hfill
  \subfloat[]{\includegraphics[width=0.2\linewidth]{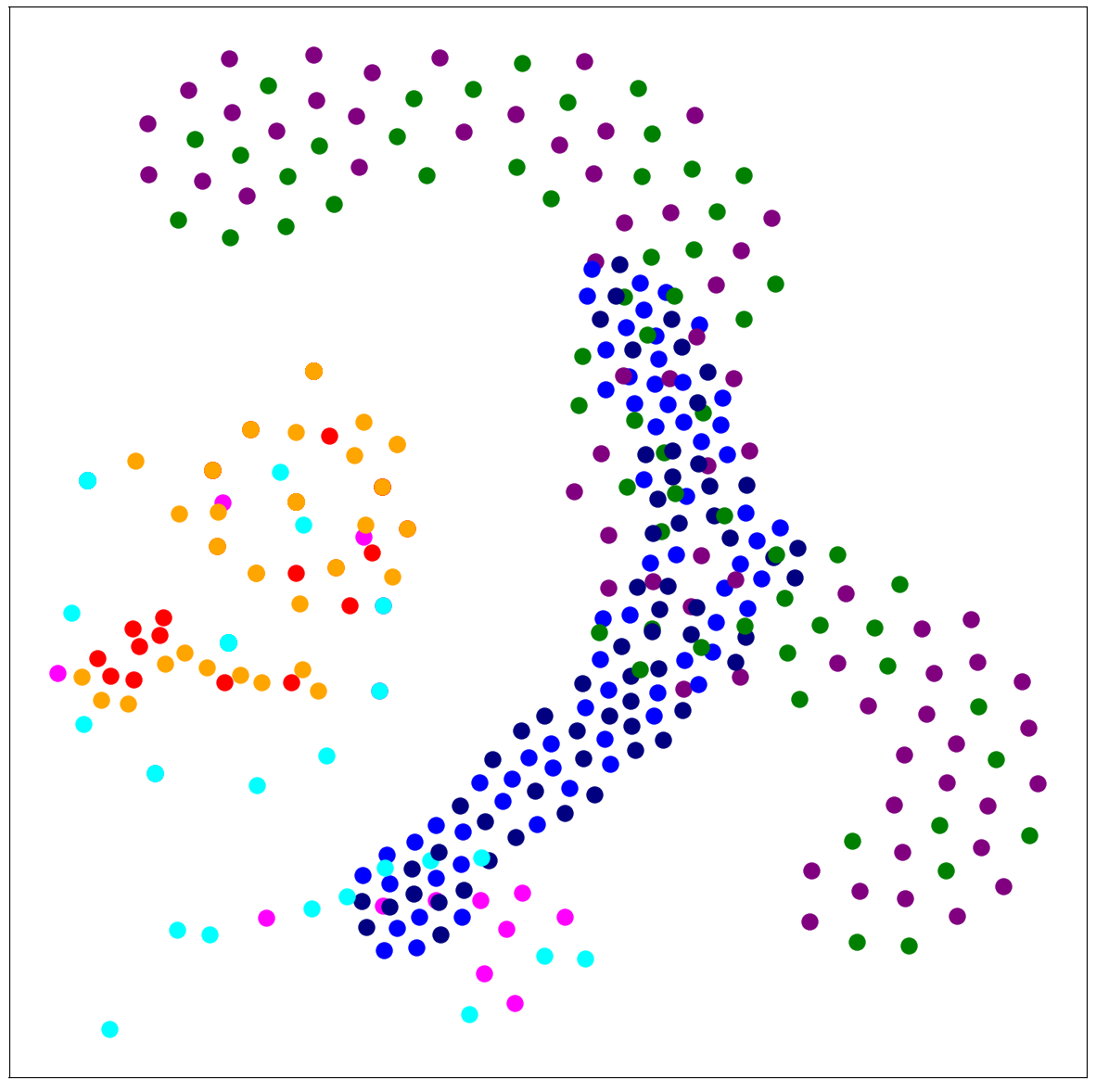}\label{fig1:d}}
   \caption{t-SNE visualizations of the decomposed unique and non-unique features. 3D features were visualized by stacking the features from slices of each subject. Red: Visual pathway (VP) with T1-weight unique features; Cyan: VP with T1-weight non-unique features; Orange: Visual pathway (VP) with FA unique features; Magenta: Visual pathway (VP) with FA non-unique features; Blue: Background with T1-weight non-unique features; Purple: Background with T1-weight unique features; Green: Background with FA unique features; Navy: Background with FA non-unique features. Subplots (a) and (c) show the unique features, revealing a distinct boundary between the VP and the background. This distinct separation indicates that using unique features in the final delineation stage allows the network to accurately isolate the VP. In contrast, once the non-unique features are introduced (subplots (b) and (d)), the feature points become intertwined in the t-SNE plots, making it more challenging to distinguish between the VP and the background.}
    \label{figure1_1}
\end{figure*}

To tackle these challenges, we draw inspiration from feature decomposition methods used in multi-modal natural image processing \cite{Deng2023} and semi-supervised learning techniques for medical image delineation \cite{wang2021annotation, tarvainen2017mean}. In this study, we propose a novel semi-supervised multi-parametric feature decomposition framework for visual pathway (VP) delineation. To address the complex inter-sequence relationships inherent in multi-parametric MRI data, we introduce a Correlation-Constrained Feature Decomposition (CFD) module. This module decomposes each MRI sequence into "unique" and "non-unique" features, which contribute to model optimization at different stages of the network.
Non-unique features assist in optimizing the network by aligning features from T1-weighted and FA images, while unique features are directly incorporated into the final segmentation stages for accurate visual pathway extraction. As demonstrated in Fig. \ref{fig1:a}\&\ref{fig1:c}, utilizing only the unique features enables the network to clearly distinguish the visual pathway from the background, resulting in well-defined boundaries. However, when both unique and non-unique features are used together (see Fig. \ref{fig1:b}\&\ref{fig1:d}, the feature points become entangled, making it more challenging to separate the visual pathway from the background. This observation underscores the critical role of feature decomposition in achieving accurate segmentation. Furthermore, we develop a consistency-based sample enhancement (CSE) module to achieve enhanced semi-supervised VP delineation performance by exploiting unlabeled data better. CSE effectively generates and promotes consistent unlabeled samples for training based on an inconsistency score. Overall, the main contributions of this work can be summarized as follows:

\begin{itemize}
    \item We design a correlation-constrained feature decomposition (CFD) that decomposes each MRI sequence into unique and non-unique features. The non-unique features contribute to the feature alignment, whereas the unique features are more crucial for fine-grained visual pathway delineation in the final stages of the network. This allows us to clearly distinguish the visual pathway from the background, with a well-defined boundary.
    \item We introduce a labeling augmentation mechanism with the designed consistency-based sample enhancement (CSE) module to address the limited availability of annotated data. This mechanism leverages the unlabeled data by generating and promoting consistent samples that exhibit consistent delineation results. These generated samples are then used to augment the labeled data, enhancing the overall delineation performance and reducing the reliance on large amounts of annotations.
    \item We evaluate the performance of the proposed approach on the HCP, multimodal brain MRI, and MDM datasets. The results obtained validate the effectiveness of our approach.
\end{itemize}


\section{Proposed Method} \label{method}
\begin{figure*}[h]
\centering 
\includegraphics[width=0.8\linewidth]{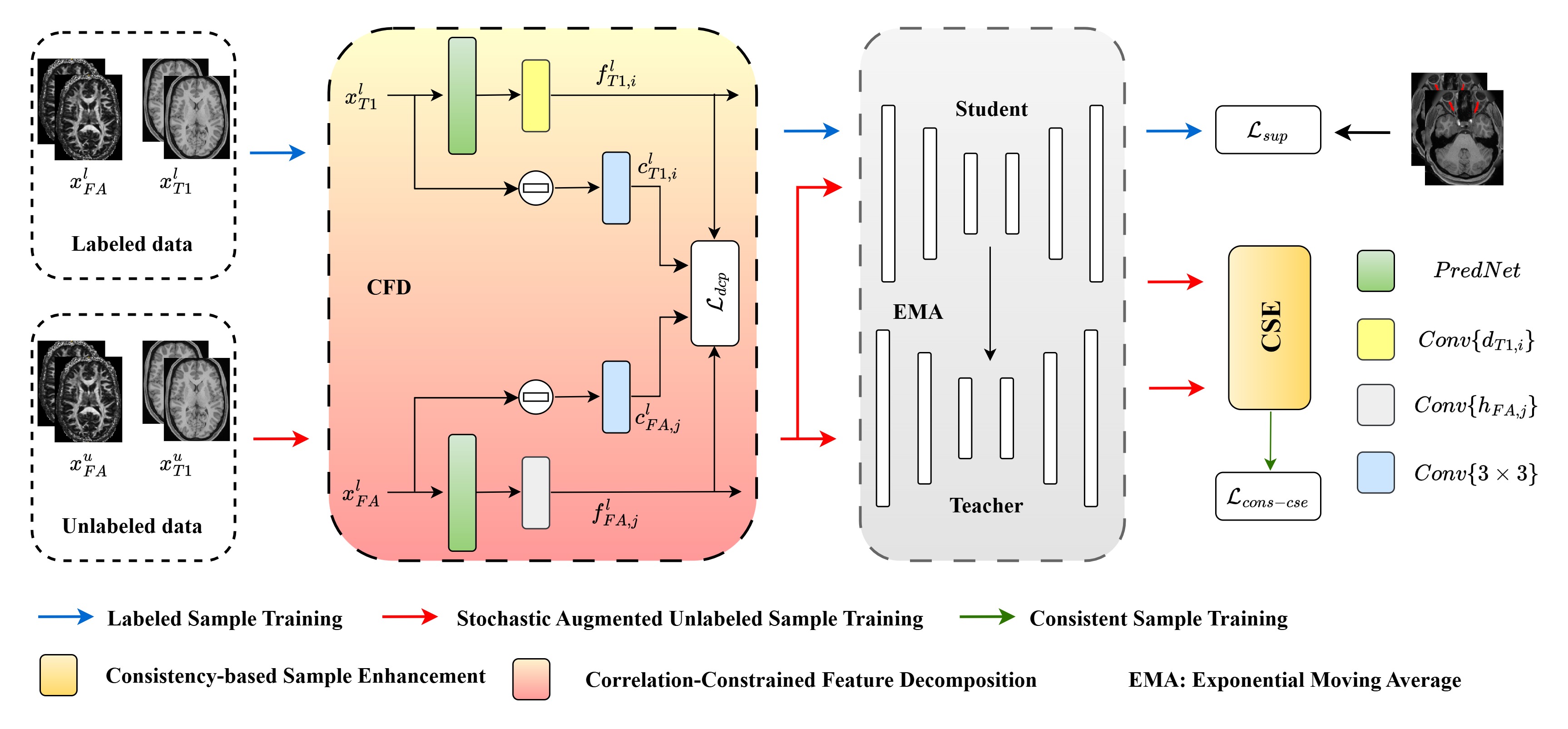}
\caption{The overall pipeline of the proposed framework. It consists of a CFD module to decompose MRI sequences and a CSE module to generate consistent samples from unlabeled data to augment labeled data. $\mathcal{L}_{sup}$ and $\mathcal{L}_{cons\_cse}$ are the supervised and unsupervised consistency loss, respectively. Here, we only presented the CFD module for the labeled images for simplicity, as the process is the same for unlabeled images.}
\label{fig2}
\end{figure*}

Let $\textbf{x}_{T1}^l$ and $\textbf{x}_{FA}^l$ be the labeled images from  the T1-weighted and FA sequence respectively, where $l$ means the image is labeled. Let $\textbf{y}$  denote the VP label. Then, their counterpart unlabeled images can be denoted by $\textbf{x}_{T1}^u$ and $\textbf{x}_{FA}^u$, where $u$ means the image is unlabeled. The proposed framework aims to delineate the VP label $\textbf{y}$ based on the images of the T1-weighted and FA sequences. To achieve this goal, the proposed framework consists of a correlation-constrained feature decomposition (CFD) and a delineation network with a consistency-based sample enhancement (CSE) module (see Fig.\ref{fig2}). The unique and non-unique features produced by CFD using from $\textbf{x}_{T1}^l$ and $\textbf{x}_{FA}^l$ are denoted by $\{\textbf{f}_{T1, i}^l, \textbf{c}_{T1, i}^l\}_{i=1}^{nP}$ and $\{\textbf{f}_{FA, j}^l, \textbf{c}_{FA, j}^l\}_{j=1}^{nQ}$, respectively. $nP$ and $nQ$ are the number of unique and non-unique feature samples from $\textbf{x}_{T1}^l$ and $\textbf{x}_{FA}^l$, respectively. Similarly, their counterpart features produced by CFD using $\textbf{x}_{T1}^u$, $\textbf{x}_{FA}^u$ are denoted by $\{\textbf{f}_{T1, i}^u$, $\textbf{c}_{T1, i}^u\}_{i=1}^{nP}$ and $\{\textbf{f}_{FA, j}^u, \textbf{c}_{FA, j}^u\}_{j=1}^{nQ}$, respectively. The delineation network contains two branches, a student model and a teacher model. The unique features $\textbf{f}_{T1, i}^l$ and $\textbf{f}_{FA, j}^l$ are fed to the student model to calculate the supervised loss. The unique features {$\textbf{f}_{T1, i}^u$ and $\textbf{f}_{FA, j}^u$} and their $m_{th}$ augmented versions $\textbf{f}_{T1, i}^{mu}$ and $\textbf{f}_{FA, j}^{mu}$ are fed to the student and teacher models, respectively to compute the unsupervised loss with CSE, where $m$ ranges from $1$ to $\textbf{M}$. 

\begin{table}[!htb]
  \caption{List of main notations used in this study}
  \resizebox{\linewidth}{!}{%
  \begin{tabular}{|c|c|}
    \hline
    Notation & Description \\
    \hline
    $\textbf{x}_{T1}^l$ & The labeled input image of T1-weighted sequence \\
    $\textbf{x}_{FA}^l$ & The labeled input image of FA sequence \\
    $\textbf{x}_{T1}^u$ & The unlabeled input image of  T1-weighted sequence \\
    $\textbf{x}_{FA}^u$ & The unlabeled input image of FA sequence \\
    $\textbf{y}$ & The VP ground truth \\
    $\textbf{f}_{T1, i}^l$ & The $i_{th}$ labeled unique feature of image  $\textbf{x}_{T1}^l$ \\
    $\textbf{f}_{FA, j}^l$ & The $j_{th}$ labeled unique feature of image  $\textbf{x}_{FA}^l$ \\
    $\textbf{c}_{T1, i}^l$ & The $i_{th}$ labeled non-unique feature of image  $\textbf{x}_{T1}^l$ \\
    $\textbf{c}_{FA, j}^l$ & The $j_{th}$ labeled non-unique feature of image  $\textbf{x}_{FA}^l$ \\
    $\textbf{c}_{T1, i}^u$ & The $i_{th}$ unlabeled non-unique feature of image  $\textbf{x}_{T1}^u$ \\
    $\textbf{c}_{FA, j}^u$ & The $j_{th}$ unlabeled non-unique feature of image  $\textbf{x}_{FA}^u$ \\
    $\textbf{f}_{T1, i}^u$ & The $i_{th}$ unlabeled unique feature of image  $\textbf{x}_{T1}^u$ \\
    $\textbf{f}_{FA, j}^u$ & The $j_{th}$ unlabeled unique feature of image  $\textbf{x}_{FA}^u$ \\
    $\textbf{f}_{T1, i}^{mu}$ & The $m_{th}$ augmented version of $\textbf{f}_{T1, i}^u$ \\
    $\textbf{f}_{FA, j}^{mu}$ & The $m_{th}$ augmented version  $\textbf{f}_{FA, j}^u$ \\
    $\textbf{f}_{T1, i}^{u_1}$ & The consistent unlabeled unique features from $\textbf{f}_{T1, i}^{u}$ \\
    $\textbf{f}_{FA, j}^{u_1}$ & The consistent unlabeled unique features from $\textbf{f}_{FA, j}^{u}$ \\
    $\textbf{d}_{T1, i}^l$ & The $i_{th}$ labeled unique filter of image  $\textbf{x}_{T1}^l$ \\
    $\textbf{h}_{FA, j}^l$ & The $j_{th}$ labeled unique filter of image  $\textbf{x}_{FA}^l$ \\
    $\textbf{S}$ & Student model \\
    $\textbf{T}$ & Teacher model \\
    $\theta^{S}$ & Student model's parameters\\
    $\theta^{T}$ & Teacher model's parameters\\
    $\hat{\textbf{p}}^m$ & The $m_{th}$ Probability vector \\ 
    $\epsilon$ & The hyper-parameter set so that the $PCC(\textbf{c}_{T1, i}^l, \textbf{c}_{FA, j}^l)$ is $+$ \\
    $\sigma$ & The regularization parameter (Eq. 3\&4) \\
    $\psi(\cdot)$ & The sparsity constraint (Eq. 3\&4) \\
    $\gamma$ & A decay factor controlling the EMA \\
    $K$ & The number of categories (here 2) \\
    $\mathcal{L}_{sup}$ & The supervised loss \\
    $\mathcal{L}_{dcp}$ & The decomposition loss \\
    $\mathcal{L}_{cons\_cse}$ & The unsupervised loss \\
    $\delta$ & The current consistency weight (Eq. 14) \\
    $\beta$ & The hyper-parameter associated with $\mathcal{L}_{dcp}$ \\
    $\alpha$ & The hyper-parameter controlling the weight of $\mathcal{L}_{sup}$ \\
    $h_{\lambda}$ & The soft-thresholding operation to induce sparsity (Fig. 3) \\
    $var(\cdot)$ & denotes the variance estimation \\
    \hline
  \end{tabular}}
  \label{tab1}
\end{table}

\subsection{VP Delineation via CFD} 
Inspired by \cite{Deng2020, Deng2023}, we modeled the dependencies between T1-weighted and FA images using multi-parametric convolutional sparse coding. Unlike similar studies, our CFD module separates MRI sequences into unique features and non-unique features. The non-unique features contribute to the feature alignment, whereas the unique features are more crucial for fine-grained visual pathway delineation in the final stages of the network.


Based on this analysis, each imaging sequence is decomposed into unique and non-unique features. We focus on the labeled images $\textbf{x}_{T1}^l$, and $\textbf{x}_{FA}^l$ for simplicity, as the decomposition process for unlabeled images $\textbf{x}_{T1}^u$, and $\textbf{x}_{FA}^u$ is identical. The dependencies between $\textbf{x}_{T1}^l$ and $\textbf{x}_{FA}^l$ are modeled as follows:

\begin{equation} \label{eq_1}
\textbf{x}_{T1}^l\ = \underbrace{\textbf{c}_{T1, i}^l}_{\textit{non-unique features}} +  \underbrace{\sum_{i}\textbf{d}_{T1, i}^l \ast \textbf{f}_{T1, i}^l}_{\textit{unique features}} 
\end{equation}
\begin{equation} \label{eq_2}
\textbf{x}_{FA}^l = \underbrace{\textbf{c}_{FA, j}^l}_{\textit{non-unique features}} +  \underbrace{\sum_{j}\textbf{h}_{FA, j}^l \ast \textbf{f}_{FA, j}^l}_{\textit{unique features}}  
\end{equation}

where the symbol $\ast$ represents the convolutional operation; $\textbf{d}_{T1, i}^l$ and $\textbf{h}_{FA, j}^l$ are the unique filters, respectively. The unique filters have their unique feature responses $\textbf{f}_{T1, i}^l$ and $\textbf{f}_{FA, j}^l$. 

Assuming all dictionary filters are known, we can calculate the unique feature responses $\textbf{f}_{T1, i}^l$ and $\textbf{f}_{FA, j}^l$ by solving the following optimization problems:
\begin{equation} \label{eq_3}
\arg \min_{\{\textbf{f}_{T1, i}^l\}} \frac{1}{2} \| \textbf{x}_{T1}^l - \textbf{c}_{T1, i}^l - \sum_{i}\textbf{d}_{T1, i}^l \ast \textbf{f}_{T1, i}^l \| + \sigma \psi(\textbf{f}_{T1, i}^l)
\end{equation}
\begin{equation} \label{eq_4}
\arg \min_{\{\textbf{f}_{FA, j}^l\}} \frac{1}{2} \| \textbf{x}_{FA}^l - \textbf{c}_{FA, j}^l - \sum_{j}\textbf{h}_{FA, j}^l \ast \textbf{f}_{FA, j}^l \| + \sigma \psi(\textbf{f}_{FA, j}^l)
\end{equation}

Where $\sigma$ is the regularization parameter.

\begin{figure}[t]

\centerline{\includegraphics[width=1.0\columnwidth]{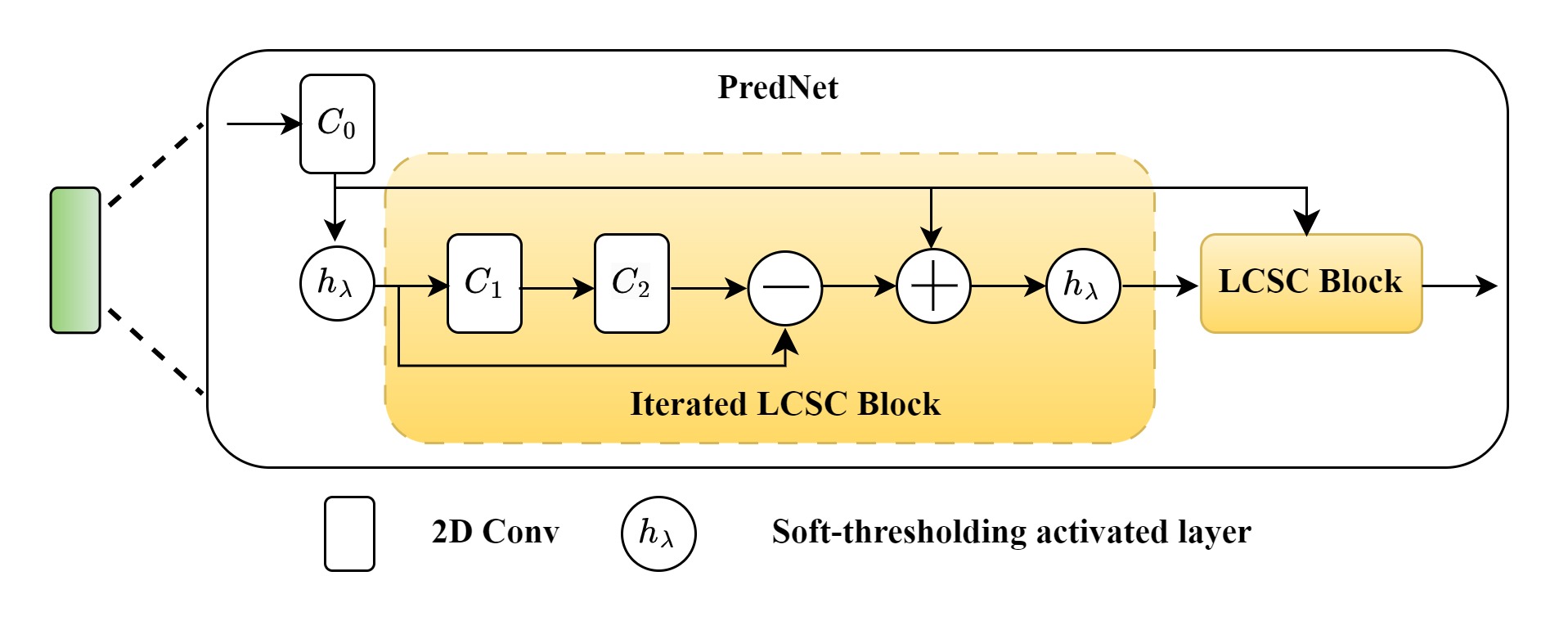}}
\caption{The detailed structure of the feature prediction block (PredNet). $C_0$, $C_1$, and $C_2$ represent the convolution layers; and $h_\lambda$ denotes the soft-thresholding operation to induce sparsity.}
\label{fig3}
\end{figure}

The decomposition is performed using multi-parametric convolutional sparse coding. Two prediction networks \cite{Deng2023} (see Fig. \ref{fig3}) are used: one for predicting the unique features $\textbf{f}_{T1, i}^l$ of $\textbf{x}_{T1}^l$, and another for predicting the unique features $\textbf{f}_{FA, j}^l$ of $\textbf{x}_{FA}^l$. Their detailed structure can be found in this study \cite{Deng2023}. The unique feature prediction block for $\textbf{x}_{T1}^l$ aims to minimize the following equation:

\begin{equation} \label{eq_5}
\arg \min_{\{\textbf{f}_{T1, i}^l\}} \frac{1}{2} \| \hat{\textbf{x}}_{T1}^l - \sum_{i}\textbf{d}_{T1, i}^l \ast \textbf{f}_{T1, i}^l \| + \sigma \psi(\textbf{f}_{T1, i}^l)
\end{equation}

where $\hat{\textbf{x}}_{T1}^l = \textbf{x}_{T1}^l - Conv(\textbf{c}_{T1, i}^l)$ represents the difference between $\textbf{x}_{T1}^l$ and the convolution of the non-unique features $\textbf{c}_{T1, i}^l$. Eq. \ref{eq_5} becomes a standard convolutional sparse coding problem, which is solvable using the learned convolutional sparse coding algorithm (LCSC) introduced in \cite{Sreter2018} when the sparsity constraint $\psi(\cdot)$ is $l_1$ norm. This gives us the network solution of $\textbf{f}_{T1, i}^l$ and forms the unique feature prediction block of $\textbf{x}_{T1}^l$.

Similarly, the unique feature prediction block for $\textbf{x}_{FA}^l$ aims to minimize the following equation:

\begin{equation} \label{eq_6}
\arg \min_{\{\textbf{f}_{FA, j}^l\}} \frac{1}{2} \| \hat{\textbf{x}_{FA}^l} - \sum_{j}\textbf{h}_{FA, j}^l \ast \textbf{f}_{FA, j}^l \| + \sigma \psi(\textbf{f}_{FA, j}^l)
\end{equation}

where $\hat{\textbf{x}}_{FA}^l = \textbf{x}_{FA}^l - Conv(\textbf{c}_{FA, j}^l)$ represents the difference between $\textbf{x}_{FA}^l$ and the convolution of the non-unique features $\textbf{c}_{FA, j}^l$.

After obtaining the unique features from each MRI sequence, the non-unique features $\textbf{c}_{T1, i}^l$ and $\textbf{c}_{FA, j}^l$ of $\textbf{x}_{T1}^l$ and $\textbf{x}_{FA}^l$, respectively, are obtained by subtracting the unique features from the original images. A decomposition loss function based on the Pearson correlation coefficient is then minimized to encourage correlation between the non-unique MRI sequence's features while reducing correlation between the unique MRI sequence's features:

\begin{equation} \label{eq_7}
\mathcal{L}_{dcp} = \frac{PCC(\textbf{f}_{T1, i}^l, \textbf{f}_{FA, j}^l)^2}{\epsilon + PCC(\textbf{c}_{T1, i}^l, \textbf{c}_{FA, j}^l)}
\end{equation}

where $PCC$ denotes the Pearson correlation coefficient, and $\epsilon$ is a hyper-parameter set to ensure the denominator is always positive.

By minimizing this decomposition loss function, the method aims to model the dependencies between the two types of images, separating the unique and non-unique features and capturing their correlations appropriately. Then, the unique features are used as inputs to the final stage of our delineation network reducing our model's susceptibility to noise signals unrelated to the visual pathway. The supervised loss function to optimize the labeled image decomposition and student model parameter sets in an end-to-end manner is a composite of the binary cross-entropy loss and the dice loss. Then, the supervised loss is estimated as follows:
\begin{equation} \label{eq_8}
\begin{split}
 \mathcal{L}_{sup} = (- \sum_{k=1}^{K} \mathbb{E}_{[\textbf{y}=k]} log (\theta^{S}(\textbf{f}_{T1, i}^l, \textbf{f}_{FA, j}^l)_{(k)}))\\
 + (1 - \sum_{k=1}^{K} \mathbb{E}_{[\textbf{y}=k]}) [\frac{2\times\theta^{S}(\textbf{f}_{T1, i}^l, \textbf{f}_{FA, j}^l)}{\|\textbf{y}\|_1 + \|\theta^{S}(\textbf{f}_{T1, i}^l, \textbf{f}_{FA, j}^l)}]_{(k)}
 \end{split}
\end{equation}

where $\mathbb{E}$ represents an indicator function, $K$ is the number of categories, and $\textbf{y}$ represents the ground truth.

\subsection{Self-Ensembling Mean Teacher With CSE}
Although the proposed VP delineation network with CFD enables learning from labeled training data, the limited number of labeled scans still hinders the model's performance. To mitigate this issue, we employ a self-ensembling mean teacher architecture to leverage unlabeled images. The teacher model $\theta^{T}$ acts as an Exponential Moving Average (EMA) of the student model $\theta^{S}$. In each training step, the teacher model's parameters are updated as follows:

\begin{equation}  \label{eq_9}
\theta_{t}^{T} = \gamma \theta_{t-1}^{T} + (1 - \gamma) \theta_{t}^{S}
\end{equation}
where $\gamma \in [0, 1]$ is a decay factor controlling the EMA, $\theta_{t-1}^{T}$ is the historical aggregation of the teacher model, and $\theta_{t}^{S}$ represents the weight of the student model. This strategy allows the teacher model to guide the student model, providing more stable and reliable predictions and reducing the impact of noisy or inconsistent samples during training. An unsupervised consistency loss is defined to measure the agreement between the predictions of the student and teacher models for the inputs $\textbf{f}_{T1, i}^u$, and $\textbf{f}_{FA, j}^u$, considering different random perturbations:

\begin{equation} \label{eq_10}
 \mathcal{L}_{cons} = \sum_{k=1}^{K} \| \theta^{S}(\textbf{f}_{T1, i}^u, \textbf{f}_{FA, j}^u)_{(k)} - \theta^{T}(\textbf{f}_{T1, i}^u, \textbf{f}_{FA, j}^u)_{(k)} \|^2 
\end{equation}

Although the self-ensembling mean-teacher model encourages the production of similar outputs for similar inputs, its performance degrades when the pseudo-labels are inconsistent \cite{qiu2025semi}. These inconsistent or unreliable pseudo-labels are likely to accumulate during semi-supervised training, leading to performance degradation \cite{yang2022st++}. Therefore, to mitigate the potential decline in performance caused by unreliable pseudo-labels, we propose a consistency-based sample enhancement (CSE) module. This module prioritizes the predictions of more reliable samples to generate higher-quality pseudo-labels. Specifically, we augmented the input images by adding Gaussian noise (with a mean of zero and a standard deviation of 0.1). By passing the augmented images through the CFD module, we obtained the corresponding augmented unique features ($\textbf{f}_{T1, i}^{mu}$ and $\textbf{f}_{FA, j}^{mu}$). Then, their corresponding probability vector $\hat{\textbf{p}}^m$  can be generated by feeding $\textbf{f}_{T1, i}^{mu}$ and $\textbf{f}_{FA, j}^{mu}$ into the student and teacher models. Furthermore, the consistency of the predictions on $\textbf{f}_{T1, i}^{mu}$ and $\textbf{f}_{FA, j}^{mu}$ by the student and teacher models can be measured by the standard deviation of these $\textbf{M}$ probability vectors $ [\hat{\textbf{p}}^{1}, \hat{\textbf{p}}^{2}, \ldots, \hat{\textbf{p}}^{M}]$, which is calculated as:

\begin{equation} \label{eq_11}
cons_{\theta^{S}} = \sum_{k=1}^{K}\sqrt{var\Bigl(\Bigl[\hat{\textbf{p}}_k^{1}, \hat{\textbf{p}}_k^{2}, \ldots, \hat{\textbf{p}}_k^{M}\Bigl]\Bigl)}
\end{equation}

\begin{equation} \label{eq_12}
cons_{\theta^{T}} = \sum_{k=1}^{K}\sqrt{var\Bigl(\Bigl[\hat{\textbf{p}}_k^{1}, \hat{\textbf{p}}_k^{2}, \ldots, \hat{\textbf{p}}_k^{M}\Bigl]\Bigl)}
\end{equation}

where $cons_{\theta^{S}}$ and $cons_{\theta^{T}}$ denote the consistency of the predictions from the student and teacher models, respectively. 

Moreover, the inconsistency score is calculated by computing the mean squared error between $cons_{\theta^{S}}$ and $cons_{\theta^{T}}$ for each unlabeled sample. Besides, we apply a warm-up function ($1 - \frac{e}{e_{max}}$) to the inconsistency score, gradually increasing it during the warm-up training period, where $e$ and $e_{max}$ are the current and total number of epochs, respectively. Finally, to generate higher-quality pseudo-labels, we prioritize the predictions of more reliable samples  (inconsistency score less than a threshold value set at $0.05$) to calculate the unsupervised consistency loss $\mathcal{L}_{cons\_cse}$, defined as follows:

\begin{equation} \label{eq_13}
 \mathcal{L}_{cons\_cse} = \sum_{k=1}^{K} \| \theta^{S}(\textbf{f}_{T1, i}^{u_1}, \textbf{f}_{FA, j}^{u_1})_{(k)} - \theta^{T}(\textbf{f}_{T1, i}^{u_1}, \textbf{f}_{FA, j}^{u_1})_{(k)} \|^{2} 
\end{equation}
where $\textbf{f}_{T1, i}^{u_1}$ and $\textbf{f}_{FA, j}^{u_1}$ represent the generated consistent unlabeled data for the T1-weighted and FA sequences, respectively.

The combination of the self-ensembling mean teacher model and the $CSE$ module enhances the overall effectiveness of our semi-supervised multi-parametric VP delineation framework. By leveraging the stability of the teacher model and generating consistent samples for training, our approach reduces the dependency on large labeled datasets and improves the accuracy of VP delineation.

The full objective function for training our model can be formulated as follows:

\begin{equation} \label{eq_14}
\arg \min_{\theta^{S}} (\alpha \mathcal{L}_{sup} + \delta \mathcal{L}_{cons\_cse} + \beta \mathcal{L}_{dcp})
\end{equation}

Here, $\alpha$ is a hyper-parameter controlling the weight of $\mathcal{L}_{sup}$, $\delta$ represents the current consistency weight of $\mathcal{L}_{cons\_cse}$, and $\beta$ is a hyper-parameter associated with $\mathcal{L}_{dcp}$.

\section{Experimental Setup} \label{exp}
\subsection{Dataset Acquisition}

Two datasets were used in this study, including the Human Connectome Project (HCP) dataset \cite{van2012human}, the Multi-shell Diffusion MRI (MDM) dataset \cite{tong2019reproducibility}, and the multimodal brain MRI dataset \cite{rezende2019test}. This study used 82 HCP data (16 cases labeled and 66 cases unlabeled) and 8 MDM (3 cases labeled and 5 cases unlabeled) for training. Furthermore, 10 HCP cases and 2 MDM cases were employed to evaluate the proposed method. The data preprocessing and VP ground truth generation steps can be found in this study \cite{Xie2023}. In addition, the FA images from both datasets were obtained using the Dipy package.

\subsubsection{The HCP dataset} The HCP dataset \cite{van2012human} offers high-quality dMRI and T1-weighted (T1w) data, which the local Institutional Review Board of Washington University has approved. The dMRI acquisition parameters for the HCP dataset are as follows: TR = $5520\si{\milli\second}$, TE = $89.5\si{\milli\second}$, FA = 78°, voxel size = $1.25 \times 1.25 \times 1.25\si{\milli\metre\cubed}$, FOV = $210 \times 180\si{\second\per_/\meter\squared}$, 18 base images with $b-values = 5 \si{\second\per_/\meter\squared}$), and 270 gradient directions with three other b-values (1000, 2000, and 3000 $\si{\second\per_/\meter\squared}$). Additionally, the T1w acquisition parameters for the HCP dataset are as follows: TR = $2400\si{\milli\second}$, TE = $2.14\si{\milli\second}$, and voxel size = $1.25 \times 1.25 \times 1.25\si{\milli\metre\cubed}$. 

\subsubsection{MDM dataset} The MDM dataset \cite{tong2019reproducibility} was obtained from three traveling subjects using consistent acquisition settings across ten imaging centers. The data were acquired using a Simultaneous Multi-Slice EPI sequence with an isotropic spatial resolution of $1.5\si{\milli\metre\cubed}$. The dMRI acquisition parameters for the MDM dataset are as follows: six base images with b-values = 0 $\si{\second\per_/\meter\squared}$, 90 gradient directions with three other b-values (1000, 2000, and 3000 $\si{\second\per_/\meter\squared}$), and voxel size = $1.5 \times 1.5 \times 1.5\si{\milli\metre\cubed}$. Additionally, the T1w acquisition parameters for the MDM dataset are as follows: voxel size = $1.2 \times 1.0 \times 1.0\si{\milli\metre\cubed}$.

\subsubsection{Multimodality brain MRI} . The multimodality brain MRI dataset \cite{rezende2019test} comprises test-retest MRIs of 45 healthy participants (13 males, aged 22 to 35 years), acquired using 3T MRI machines at varying intervals (mean interval of 4.7 ± 2 months, with a minimum of 1 month and a maximum of 11 months). It is worth noting that two participants do not have available diffusion-weighted imaging (DWI) for the retest. A brief overview of the imaging protocol is provided below: (a) T1‐WI: axial orientation, FOV = 224 × 224 $\si{\milli\metre}$, voxel size 0.7 $\si{\milli\metre\cubed}$ (isotropic), TR/TE/TI 2400/2.14/1000 $\si{\milli\second}$, flip angle 8°; (b) DWI: 18 b0 images, 90 gradient directions, b = 3,000 $\si{\second\per_/\meter\squared}$), TE/TR 89/5520 $\si{\milli\second}$, 1.25 $\si{\milli\metre}$ voxel (isotropic); (c) rsfMRI: EPI sequence, voxel 2 $\si{\milli\metre\cubed}$ (isotropic), TR/TE 720/33.1 $\si{\milli\second}$, flip angle 52°, 72 slices, 1200 frames per run.

\subsection{Implementation Details}
\subsubsection{Network Structure} 
The structure of the feature prediction block consists of a learnable convolutional layer, a soft-thresholding operator $h_{\lambda}$, and $n$ number of LCSC block (here $n=2$) and was adopted from this work \cite{Deng2023}. We adopt a 2D U-shaped encoder-decoder architecture similar to \cite{Li2019} with skip connections between encoding and decoding layers for the delineation module structure. Channel-wise concatenation is employed instead of pixel-wise summation, reducing network weight by halving convolution kernels in the encoder with max-pooling. A spatial attention block is introduced to enable the T1 weighted-relevant features to generate supervision information that influences the learning of the FA-relevant features. Skip connections in the decoder retain spatial information lost during decoding. 
\subsubsection{Training Details}
During the training phase, we employed an initial learning rate of 0.0002 and a weight decay of 0.00001. All experiments were conducted on a workstation equipped with NVIDIA RTX GPUs (24GB). We employed various data augmentation techniques to enhance the training process, such as level flipping and random color attribute changes, and the models were trained for 200 epochs. Additionally, the data were resized to $128 \times 160 \times 128$ and $128 \times 128 \times 92$, respectively. We set $\alpha = 10$, $\beta=1$, $M=3$, and threshold value $thres=0.05$ as optimal values for hyper-parameters obtained through extensive experiments. 


\section{Results}
\subsection{Comparison with Existing Methods}
\hfill\\
To demonstrate the effectiveness of our approach in utilizing additional supervision from unlabeled and multi-parametric MRI data, we conducted extensive comparisons with several semi-supervised learning (SSL) methods and three fully supervised (FS) methods. 

We began by contrasting our method with a Baseline model where a 2D U-Net is used as lower-bound and trained using 8 and 16 annotated samples for the HCP dataset, and 1 and 3 annotated samples for the MDM dataset. Subsequently, we compared our approach with advanced SSL methods, including Mean Teacher (MT) \cite{tarvainen2017mean}, Uncertainty Aware Mean Teacher (UAMT) \cite{yu2019uncertainty}, Dual-task Consistency (DTC) \cite{luo2021semi}, Uncertainty-guided Mutual Consistency Learning (UG-MCL) \cite{zhang2023uncertainty}, Bidirectional Copy-Paste (BCP) \cite{Bai2023}, and Dynamically Mixed Soft Pseudo-label Supervision (DMSPS) \cite{han2024dmsps}. All SSL methods rely on U-Net architectures (except DMSPS which uses UNet-CCT \cite{han2024dmsps}, a variant of U-Net). The U-Net model is structured with an encoding and a decoding segment. The encoding section employs two $3 \times 3$ convolutional layers with a stride of one, which is followed by batch normalization (BN), a rectified linear unit (ReLU), and a $2 \times 2$ max-pooling operation with a stride of two across four levels. With each downsampling step, the input image’s dimensions are halved while the number of feature channels is doubled. At the lowest level, there are two $3 \times 3$ convolutional layers accompanied by a BN, but no pooling layer is applied. The decoding section reconstructs the original dimensions of the input images by upsampling the feature map, concatenating it with the corresponding feature channels from the contracting path, and applying two $3 \times 3$ convolutional layers, followed by a BN and a ReLU. The output layer consists of a $1 \times 1$ convolution followed by a sigmoid activation function to convert the feature vector into a binary prediction. Additionally, we compared our method against FS methods, including MMFnet \cite{Xie2023_1}, TransAttUnet \cite{chen2023transattunet}, and MSDANet \cite{zhang2024msdanet}. MMFnet is used as lower-bound (Baseline) and upper-bound (FS-1), while TransAttUnet (FS-2) and MSDANet (FS-3) cater to general medical image delineation. All the comparison methods used both T1-weighted and FA images as input.

It's worth noting that all approaches were evaluated under the same partition protocols to ensure fairness, with extraction results from competing methods obtained through their official implementations. Performance metrics for each method included the Dice Score (DSC), 95\% Hausdorff Distance (HD95), and Average Surface Distance (ASD). Although our model is 2D, all the statistics are calculated in 3D by stacking the slices. Furthermore, we conducted a two-tailed paired t-test with a significance threshold of $p < 0.05$ to assess the statistical significance of the results.

\begin{center}
\begin{table*}[!h]%
\caption{Quantitative comparisons of all evaluated approaches on the HCP dataset. The top 1 results in semi-supervised learning methods are highlighted in bold.\label{tab2}}
\begin{tabular*}{\linewidth}{@{\extracolsep\fill}lllllll@{}}
\toprule
&\multicolumn{3}{@{}l}{\textbf{Labeled/Unlabeled: 16/66}} & \multicolumn{3}{@{}l}{\textbf{Labeled/Unlabeled: 8/74}} \\\cmidrule{2-4}\cmidrule{5-7}
\textbf{Methods} & \textbf{DSC $\pm$ SD} & \textbf{HD95 $\pm$ SD} & \textbf{ASD $\pm$ SD}  & {\textbf{DSC $\pm$ SD}} & \textbf{HD95 $\pm$ SD}  & \textbf{ASD $\pm$ SD}   \\
\midrule
Baseline & 0.75 $\pm$ 0.04 & 2.38 $\pm$ 1.11 & 0.88 $\pm$ 0.39 & 0.60 $\pm$ 0.05 & 6.37 $\pm$ 5.97 & 1.53  $\pm$  0.83 \\
\midrule
MT &	0.77 $\pm$ 0.04	& 15.30 $\pm$ 10.14 & 3.81 $\pm$ 2.48 &	0.75 $\pm$ 0.04 &	37.47 $\pm$ 8.73 &	15.37 $\pm$ 3.89 \\
DTC &	0.77 $\pm$ 0.07 & 1.58 $\pm$ 0.56 &	0.37 $\pm$ 0.15 &	0.72 $\pm$ 0.07 &	2.46 $\pm$ 1.92 &	0.53 $\pm$ 0.32 \\
UA-MT &	0.78 $\pm$ 0.04 & 1.61 $\pm$ 0.46 & 0.38 $\pm$ 0.13 &	0.76 $\pm$ 0.04 &	3.22 $\pm$ 1.42 &	0.83 $\pm$ 0.28 \\
UG-MCL &	0.79 $\pm$ 0.04	& 1.45 $\pm$ 0.32 & 0.32 $\pm$ 0.09 &	0.76 $\pm$ 0.05 & \textbf{1.51 $\pm$ 0.48} & 0.36 $\pm$ 0.10 \\
BCP &	0.77 $\pm$ 0.04 & 1.75 $\pm$ 0.45 & 0.39 $\pm$ 0.10 &	0.76 $\pm$ 0.05 & 1.91 $\pm$ 1.16	& 0.39 $\pm$ 0.15 \\
DMSPS &	0.78 $\pm$ 0.04 & 1.61 $\pm$ 0.46 & 0.47 $\pm$ 0.17 &	0.77 $\pm$ 0.05 & 2.09 $\pm$ 1.59	& 0.45 $\pm$ 0.25 \\
Ours &	\textbf{0.85 $\pm$ 0.03} &	\textbf{1.37 $\pm$ 0.35} &	\textbf{0.22 $\pm$ 0.06} &	\textbf{0.81 $\pm$ 0.03} &	1.62 $\pm$ 0.94 &	\textbf{0.30 $\pm$ 0.13} \\
\midrule
FS-1	& 0.87 $\pm$ 0.02 &	1.25 $\pm$ 0.00 &	0.18 $\pm$ 0.03 &	0.87 $\pm$ 0.02 &	1.25 $\pm$ 0.00 &	0.18 $\pm$ 0.03 \\

\bottomrule
\end{tabular*}
\begin{tablenotes}
\item[$^{\rm a}$] Baseline: Lower-bound baseline is trained only with the 16 (20\%) and 8 (10\%) annotated samples, respectively.  
\end{tablenotes}
\end{table*}
\end{center}

\begin{center}
\begin{table*}[!h]%
\caption{Quantitative comparisons of all evaluated approaches on the MDM dataset. The top 1 results in semi-supervised learning methods are highlighted in bold. \label{tab3}}
\begin{tabular*}{\linewidth}{@{\extracolsep\fill}lllllll@{}}
\toprule
&\multicolumn{3}{@{}l}{\textbf{Labeled/Unlabeled: 1/7}} & \multicolumn{3}{@{}l}{\textbf{Labeled/Unlabeled: 3/5}} \\\cmidrule{2-4}\cmidrule{5-7}
\textbf{Methods} & \textbf{DSC $\pm$ SD} & \textbf{HD95 $\pm$ SD} & \textbf{ASD $\pm$ SD}  & {\textbf{DSC $\pm$ SD}} & \textbf{HD95 $\pm$ SD}  & \textbf{ASD $\pm$ SD}   \\
\midrule
Baseline & 0.34 $\pm$ 0.11 & 20.21 $\pm$ 2.19 & 4.63 $\pm$ 1.25 & 0.59 $\pm$ 0.05 & 13.91 $\pm$ 1.91 & 2.12 $\pm$ 0.30 \\
\midrule
MT &	0.55 $\pm$ 0.04	& 8.16 $\pm$ 3.14 & 1.58 $\pm$ 0.44 &	0.58 $\pm$ 0.02 &	10.87 $\pm$ 6.61 &	1.77 $\pm$ 0.72 \\
DTC &	0.57 $\pm$ 0.01 &	8.66 $\pm$ 0.98 &	3.15 $\pm$ 0.59 & 0.58 $\pm$ 0.02 & 9.78 $\pm$ 5.27 &	1.70 $\pm$ 0.47 \\
UA-MT &	0.58 $\pm$ 0.04 & 3.36 $\pm$ 0.00 & 0.89 $\pm$ 0.08 &	0.59 $\pm$ 0.05 &	3.80 $\pm$ 0.45 &	0.91 $\pm$ 0.02 \\
UG-MCL &	0.61 $\pm$ 0.06	& 3.00 $\pm$ 0.00 & 0.79 $\pm$ 0.12 &	0.63 $\pm$ 0.03 & 3.18 $\pm$ 0.18 & 0.81 $\pm$ 0.01 \\
BCP &	0.59 $\pm$ 0.03 & 3.18 $\pm$ 0.18 & 0.84 $\pm$ 0.02 &	0.60 $\pm$ 0.03 & 3.52 $\pm$ 0.17	& 0.90 $\pm$ 0.01 \\
DMSPS &	0.55 $\pm$ 0.12 & 3.88 $\pm$ 0.88 & 1.05 $\pm$ 0.38 &	0.59 $\pm$ 0.05 & 3.19 $\pm$ 0.17	& 0.85 $\pm$ 0.09 \\
Ours &	\textbf{{0.70 $\pm$ 0.08}} &	\textbf{2.25 $\pm$ 0.75} &	\textbf{0.57 $\pm$ 0.20} &	\textbf{{0.74 $\pm$ 0.04}} &	\textbf{{1.82 $\pm$ 0.31}} &	\textbf{{0.43 $\pm$ 0.08}} \\
\midrule
FS-1 &	0.72 $\pm$ 0.04 & 2.13 $\pm$ 0.00  & 0.50 $\pm$ 0.02 & 0.72 $\pm$ 0.04 & 2.13 $\pm$ 0.00  & 0.50 $\pm$ 0.02 \\

\bottomrule
\end{tabular*}
\begin{tablenotes}
\item[$^{\rm a}$] MDM results are obtained by fine-tuning the different methods with their weights obtained on the HCP dataset.
\item[$^{\rm b}$] Baseline: Lower-bound baseline is trained only with the 1 (20\%) and 3 (40\%) annotated samples, respectively.
\end{tablenotes}
\end{table*}
\end{center}

\subsubsection{Performance on the HCP dataset} 
Our approach was evaluated against six state-of-the-art (SOTA) semi-supervised segmentation methods as well as four supervised baselines using the HCP dataset. The supervised FS-1 model utilizing  8 and 16 annotated samples was designated as the Baseline (lower bound) comparisons. Meanwhile, the supervised  FS-1 models utilizing 82 annotated samples were designated as the FS-1 (upper bound) comparisons. Table \ref{tab2} presents the performance metrics for all SOTA methods assessed with both 8 and 16 labeled samples, where the optimal performance values from the semi-supervised segmentation algorithms are highlighted in bold. 

As indicated in Table \ref{tab2}, all semi-supervised approaches demonstrate a greater or comparable ability to enhance segmentation outcomes compared to the supervised baseline using 8 and 16 labeled samples, underscoring the value of leveraging diverse and extensive information available in the unlabeled data for model training. Furthermore, the performance of all semi-supervised algorithms improves as the amount of labeled data increases. Notably, our method achieves significant enhancements in segmentation performance compared to other semi-supervised techniques. For instance, our approach outperforms the DTC and MT methods, achieving DSC increases of 13\% and 8\% at least for the 8 and 16 labeled data scenarios, respectively. The leading UG-MCL method consistently underperformed in all metrics when compared to ours. In addition, our method achieves statistically superior performance compared to other semi-supervised segmentation methods in terms of DSC, as confirmed by a two-tailed paired t-test ($p < 0.05$, see Fig \ref{fig7:a}). 

\subsubsection{Performance on the MDM dataset} Table \ref{tab3} presents a quantitative comparison of our method’s performance (fine-tuned with their weights obtained on the HCP dataset) when using approximately 1 and 3 subjects of the MDM dataset as labeled samples, respectively. 
We first compared our method against a Baseline (2D U-Net trained with 1 and 3 annotated samples), which is a lower benchmark for semi-supervised approaches since it does not utilize any unlabeled data. With just 1 labeled sample, our method demonstrates improvements in DSC: a 36\% increase compared to the Baseline, a remarkable 15\% increase compared to the under-performing semi-supervised approach (UA-MT), and a 9\% increase compared to the most competitive method (UG-MCL). When we increase the labeled samples to 3, the enhancements remain large compared to the Baseline and substantial compared to semi-supervised methods, reinforcing the effectiveness of our method in fully leveraging unlabeled data to understand the overall data distribution better. In addition, a statistical analysis is performed using a two-tailed paired t-test between our method and the competing methods in terms of DSC values. It can be observed in Fig \ref{fig7:b} that we obtain $p-values<0.05$, demonstrating statistically significant results over other semi-supervised segmentation techniques. 

\subsubsection{Comparison with fully supervised methods} Furthermore, we compared our method against the fully supervised methods. As observed in our experiments (see Table \ref{tab2}\&\ref{tab3}), the fully supervised methods, like FS-1, tend to depend heavily on large labeled datasets to achieve optimal accuracy \cite{qi2020small}. In contrast, our approach is designed to leverage limited labeled samples more effectively with the help of available unlabeled samples. For instance, when the dataset is very small labeled samples such as in Table \ref{tab3} and Fig \ref{fig5:b}, our method demonstrated comparable performance over FS-1, FS-2, and FS-3. This is particularly due to the fact that our approach utilizes the additional information from unlabeled data to better capture the underlying structure of the data distribution. This dual strategy leads to a more robust representation, allowing our method to effectively learn from both known and unknown samples. Moreover, we applied a reliable mean-teacher mechanism (CSE module), which further enhanced the performance by providing more informative learning signals from the unlabeled data. Conversely, it is essential to recognize that on the HCP dataset, which includes a much larger labeled set of 82 samples, FS-1, FS-2, and FS-3, achieved a higher DSC score of 0.87, 0.85, and 0.85 compared to our method's 0.85 (see Table \ref{tab2} and Fig \ref{fig5:a}. This observation indicates that while fully supervised methods can excel in rich data environments, they may not be efficient when data is scarce. Therefore, we assert that the true strength of our method is its robustness in low-label scenarios by leveraging unlabeled data, which is increasingly relevant in fields where annotation costs are high and data scarcity is common.

\begin{figure*}[!t]
\centering
\includegraphics[width=1.0\linewidth]{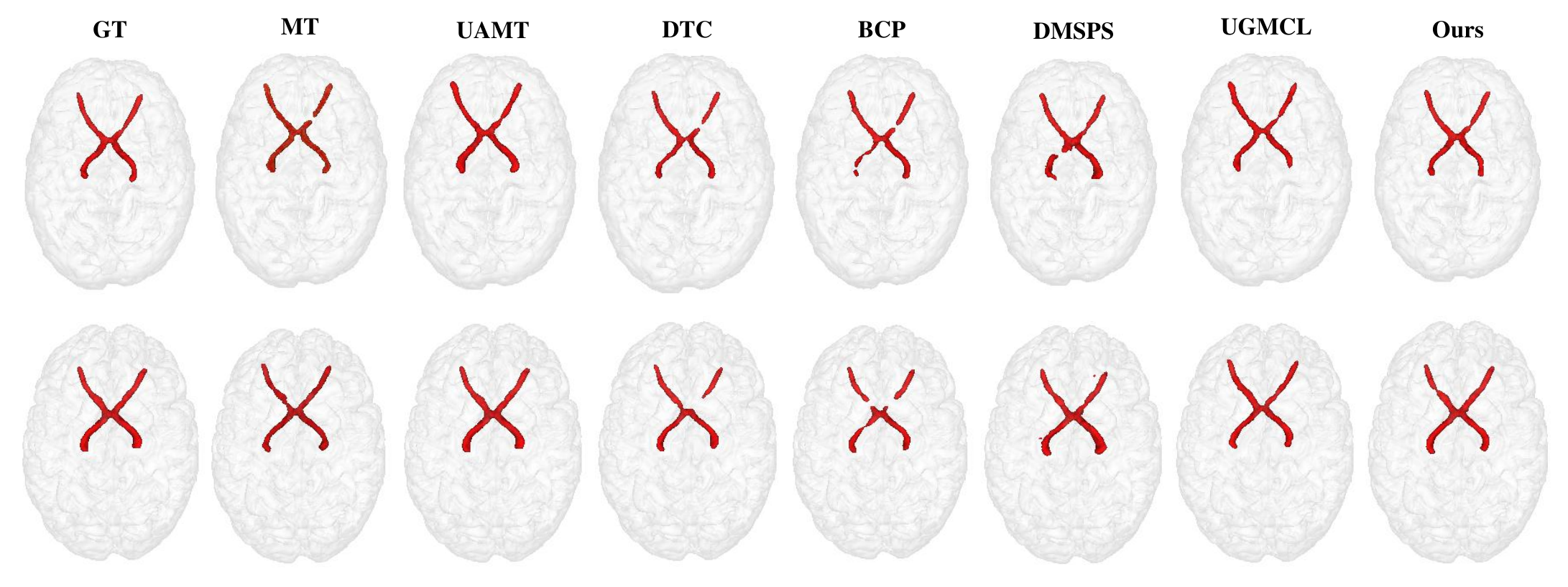} 
\caption{Qualitative results comparison on the HCP dataset. In addition to the delineation results, we provide the binary masks (ground truth) for better comparison.}
\label{figure4}
\end{figure*}

\begin{figure}[!t]
  \centering
  \subfloat[]{\includegraphics[width=0.5\linewidth]{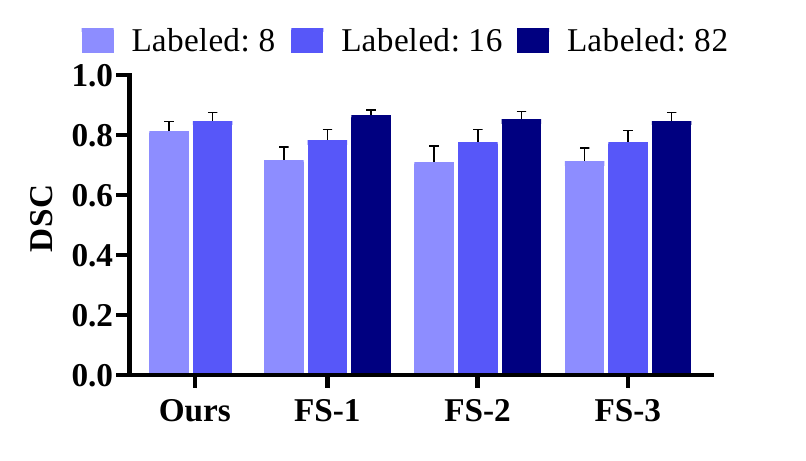}\label{fig5:a}}
  \subfloat[]{\includegraphics[width=0.5\linewidth]{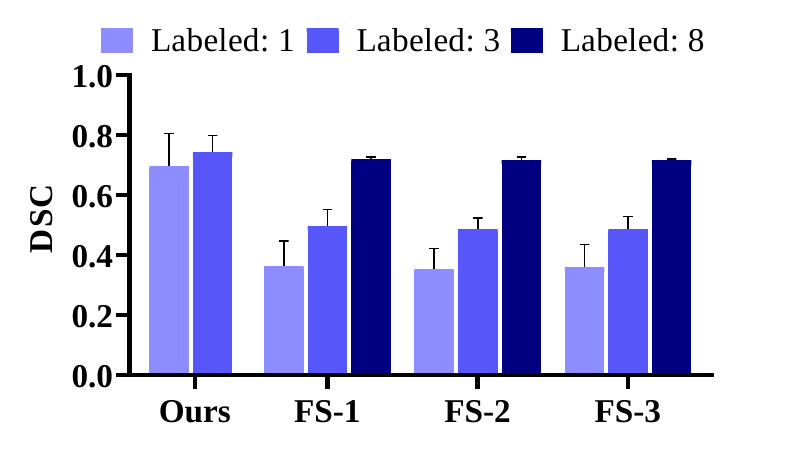}\label{fig5:b}}
  \caption{Comparison with fully supervised methods when the available labeled data is 8, 16, and 82 samples for the HCP dataset, and 1, 3, and 8 samples for the MDM dataset.}
    \label{figure5}
\end{figure}

\begin{figure*}[!t]
\centering
\includegraphics[width=1.0\linewidth]{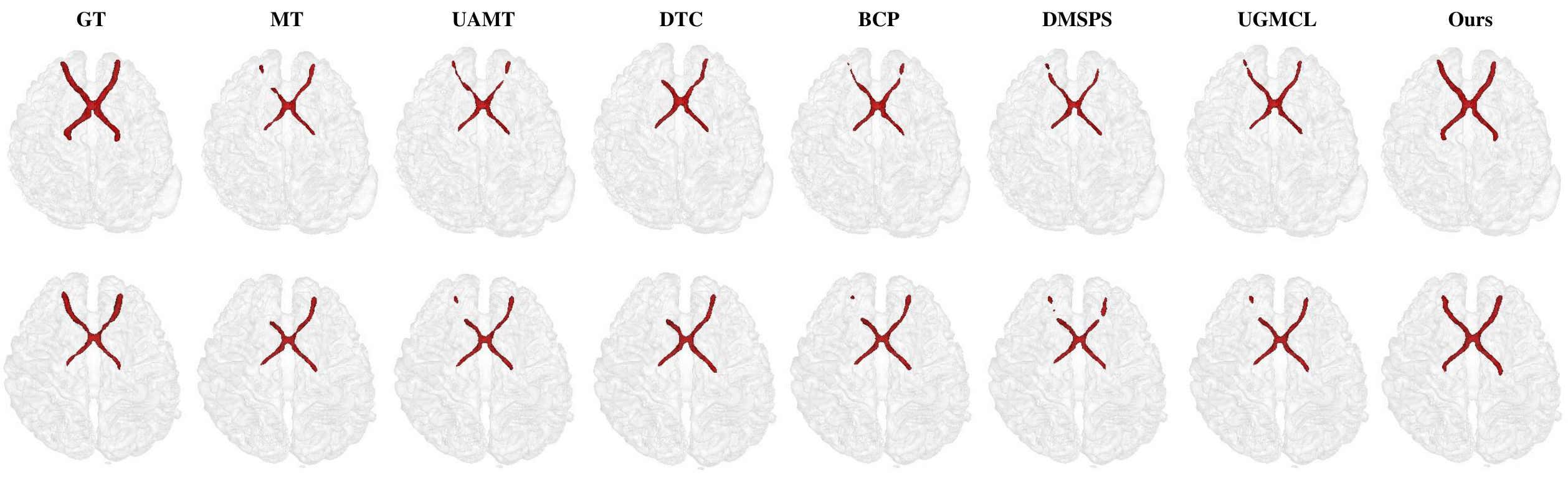} 
\caption{Qualitative results comparison on the MDM dataset. In addition to the delineation results, we provide the binary masks (ground truth) for better comparison.}
\label{figure6}
\end{figure*}

\begin{figure}[!t]
  \centering
  \subfloat[]{\includegraphics[width=0.5\linewidth]{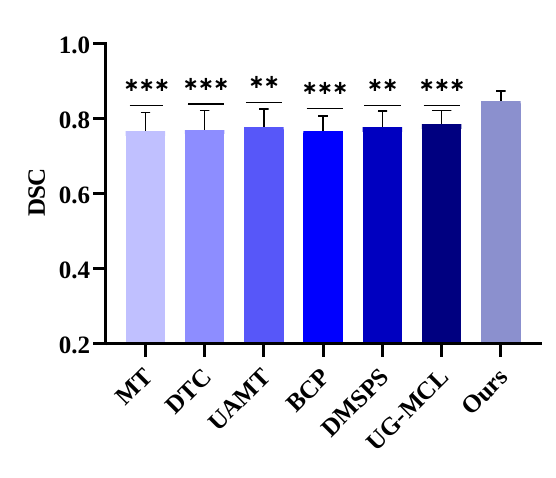}\label{fig7:a}}
  \hfill
  \subfloat[]{\includegraphics[width=0.5\linewidth]{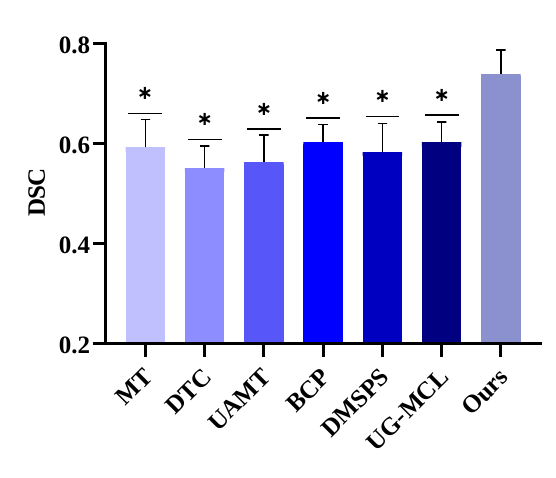}\label{fig7:b}}
   \caption{Statistical analysis between our method and the competing methods in terms of DSC values on both HCP (a) and MDM (b) datasets. An asterisk was used to signify $p < 0.05$, two asterisks for $p < 0.01$, and three asterisks for $p < 0.001$. Note that since the MDM test sample is too small, we run two Monte Carlo simulations before computing the paired t-test.}
    \label{figure7}
\end{figure}

\subsubsection{Visualization of Delineation Results}
\hfill\\
In Figs. \ref{figure4} and \ref{figure6}, we also visualize some delineation results of our proposed approach compared to other SSL methods. 

In Fig \ref{figure4}, we can see that the qualitative results of comparison methods failed to preserve the integrity of VP structures. In contrast, our method exhibits much better delineation results in all the VP structures, proving that our approach is more discriminative and consistent across structures. In Fig \ref{figure6}, we can see that MT, DTC, and UA-MT achieved the worst visualization performance compared to other methods. The  BCP, UG-MCL, and DMSPS methods achieved much better qualitative results in revealing some VP structures but showed some discontinuity near the eyeball and failed to preserve the optic tract structure. In contrast, our proposed method achieved better qualitative results in revealing the optic nerve, optic chiasm, and optic tract structures close to the ground truth.

These visualization results confirm the results presented in Tables \ref{tab2} and \ref{tab3} and demonstrate the superiority of the proposed method over the comparison methods across VP's structures and datasets.

\subsection{Ablation Study} 
We conduct a comprehensive ablation on both HCP and MDM datasets to analyze the contribution of each component and validate some hyper-parameters with 20\% and 40\% annotations, respectively. 

\subsubsection{Impact of M value}
To improve the reliability of our method, we evaluate the impact of the M value. To obtain the optimal values of $M$, we fix $\alpha=1$, $\beta=1$, and $thres = 0.05$, and then vary $M$ in $\{1, 3, 5, 7, 10\}$. The results obtained are presented in Table \ref{tab4}. The results demonstrate that the choice of $M$ value has little impact on the DSC metric, but greatly influences the ASD and HD95 metrics. Since the HD95 and ASD metrics tell us how well our model captures the exact shape and contours of VP structures, therefore, correctly selecting the $M$ value can lead to more precise VP delineation. As can be seen from Table \ref{tab4}, the optimal DSC values can be achieved when $M=10$, and optimal ASD values can be achieved when $M>1$. However, in our study, we fix $M=3$ to reduce the computation burden.

\begin{table}[!htb]
  \centering
  \caption{Impact of M value on the delineation performance}
  \begin{tabular}{cccc}
    \hline
    $M$  & DSC & HD95 (mm) & ASD (mm) \\
    \hline
    $1$  & 0.81 $\pm$ 0.03 & 1.62 $\pm$ 0.52 & 0.35 $\pm$ 0.16 \\
    $3$  & 0.85 $\pm$ 0.03 & 1.38 $\pm$ 0.38 & 0.22 $\pm$ 0.06 \\
    $5$  & 0.85 $\pm$ 0.03 & 1.38 $\pm$ 0.38 & 0.22 $\pm$ 0.07 \\
    $7$  & 0.85 $\pm$ 0.02 & 1.32 $\pm$ 0.20 & 0.23 $\pm$ 0.06 \\  
    $10$  & 0.85 $\pm$ 0.03 & 1.30 $\pm$ 0.16 & 0.22 $\pm$ 0.06 \\ 
    \hline
  \end{tabular}
  \label{tab4}
\end{table}

\subsubsection{Impact of threshold value ($thres$)}
In our study, a thresholding scheme is employed to select consistent unlabeled samples, which plays a crucial role in calculating the unsupervised consistency loss. This scheme helps ensure that only the most reliable samples are selected during the training process, enhancing the model's accuracy and robustness. As shown in Table \ref{tab5}, we explore its impact on the proposed CSE module by varying the threshold values in $\{0.05, 0.1, 0.5, 1.0\}$, and fixing $\alpha=1$, $\beta=1$, and $M=3$.  We find that the optimal results can be obtained when $thres=0.05$, and the results improve considerably when the value of $thres$ is below 0.1. This could be due to the fact that if the value of $thres$ is excessively high, the risk of overfitting to the specific subset of confident samples becomes high, making the model less performing.

\begin{table}[!htb]
  \centering
  \caption{Analysis of the impact of the thresholding scheme. The top-1 results are bolded}
  \begin{tabular}{cccc}
    \hline
    $thres$  & DSC & HD95 (mm) & ASD (mm) \\
    \hline
    $0.05$  & \textbf{0.85 $\pm$ 0.03} & 1.38 $\pm$ 0.38 & \textbf{0.22 $\pm$ 0.06} \\
    $0.1$  & 0.84 $\pm$ 0.02 & \textbf{1.25 $\pm$ 0.00} & 0.23 $\pm$ 0.05 \\
    $0.5$  & 0.84 $\pm$ 0.02 & 1.38 $\pm$ 0.38 & 0.24 $\pm$ 0.07 \\
    $1.0$  & 0.84 $\pm$ 0.03 & 1.43 $\pm$ 0.39 & 0.25 $\pm$ 0.08 \\  
    \hline
  \end{tabular}
  \label{tab5}
\end{table}

\subsubsection{Impact of Loss Weights}
\begin{figure}[!t]
  \centering
  \subfloat[]{\includegraphics[width=0.5\linewidth]{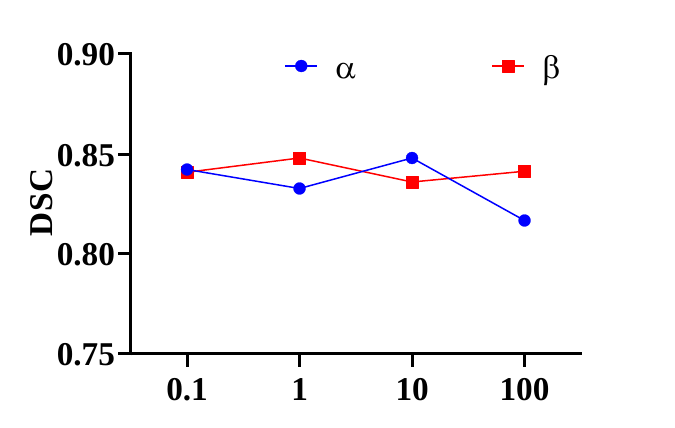}\label{fig8:a}}
  \subfloat[]{\includegraphics[width=0.5\linewidth]{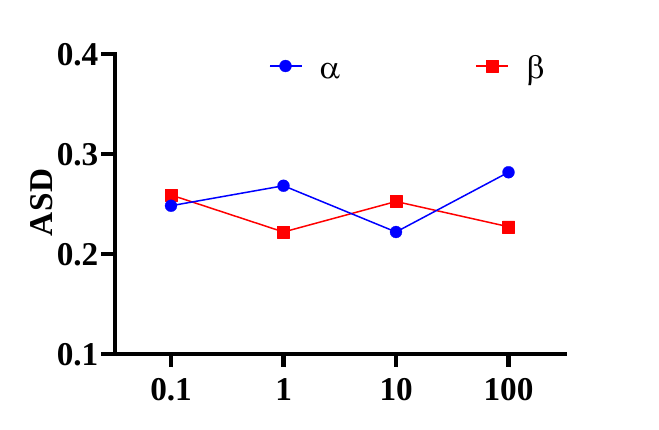}\label{fig8:b}}
   \caption{Delineation performance for the training of our model with various $\alpha$ and $\beta$ on the MDM dataset.}
    \label{figure8}
\end{figure}

The hyperparameters associated with the loss function are denoted by \(\alpha\), \(\lambda_{1}\), and \(\beta\), as defined in Equation (\ref{eq_14}). Among these, \(\lambda_{1}\) is defined as a time-dependent Gaussian function. To determine the optimal values for \(\alpha\) and \(\beta\), we apply the control variable method. Specifically, we perform a greedy search for \(\alpha\), exploring values in the following order of magnitude: \{0.1, 1, 10, 100\}. Once \(\alpha\) is fixed, we conduct a similar search for \(\beta\), also varying it across the magnitudes: \{0.1, 1, 10, 100\}. The proposed model is trained on the HCP dataset using hyperparameters \(M=3\) and \(\text{thres}=0.05\). The segmentation results with different coefficient combinations are presented in Fig. \ref{fig8:a} and \ref{fig8:b}. Analyzing the three metrics for the proposed model, we observe that optimal values for the DSC, and ASD are achieved at \(\alpha=10\) and \(\beta=1\).

\subsubsection{Effectiveness of different components}
To investigate the contribution of each component of the proposed method, we report the delineation results of different components trained with the following hyper-parameters values: $\alpha = 10$, $\beta=1$, $M=3$, and $\text{thres}=0.05$ for both HCP and MDM datasets, respectively.

As shown in Table \ref{tab6}, ‘$Model \: 1$’, ‘$Model \: 2$’, and ‘$Model \: 3$’, where ‘$Model \: 1$’ signifies the proposed method excluding the $\mathcal{L}_{dcp}$ and $CSE$ modules, ‘$Model \: 2$’ indicates the method that omits the $CSE$ module, and ‘$Model \: 3$’ corresponds to the method without $\mathcal{L}_{dcp}$. The results indicate that each component plays a crucial role in enhancing performance. Notably, the $\mathcal{L}_{dcp}$ component facilitates effective feature decomposition and increases modality/sequence diversity, leveraging Pearson correlation. When combined, ‘$CFD$’ and ‘$\mathcal{L}_{dcp}$’ in ‘$Model \: 3$’ further enhance the DSC by 1\% for the MDM dataset and 1\% for the HCP dataset. It reduces the HD95 by at least 0.22 mm for both datasets. The $CSE$ module in ‘$Model \: 2$’ capitalizes on unlabeled data, aligning the outputs of the student and teacher models effectively, which in turn boosts the DSC by an additional 3\% for the MDM dataset. Ultimately, our comprehensive approach ($ours$), utilizing all components, significantly enhances overall delineation performance, underscoring the importance of each element in our methodology.

\begin{table}[!htb]
  \centering
  \caption{Analysis of the effect of different components of the proposed approach. The top-1 results are bolded}
  \begin{tabular}{ccccccc}
    \hline
    Method  & $\mathcal{L}_{dcp}$ & $CSE$ & DSC & HD95 & ASD \\
    \hline
    HCP & & & & & \\
    \hline
    $Model \: 1$  &  & & 0.82 & 1.61 & 0.30 \\
    $Model \: 2$  & \checkmark & & 0.83 & 1.43   & 0.27 \\
    $Model \: 3$  & & \checkmark & 0.83  & 1.40  & 0.27  \\ 
    $ours$   & \checkmark & \checkmark & \textbf{0.85} & \textbf{1.37} & \textbf{0.22} \\
    \hline
    MDM & & & & & & \\
    \hline
    $Model \: 1$ &  & & 0.68 & 2.60 & 0.70 \\
    $Model \: 2$ & \checkmark & & 0.72 & 1.82 & 0.49 \\
    $Model \: 3$ &  & \checkmark & 0.69 & 2.43 & 0.61 \\  
    $ours$  & \checkmark & \checkmark & \textbf{0.74} & \textbf{1.82} & \textbf{0.43} \\
    \hline
  \end{tabular}
  \label{tab6}
\end{table}

\subsubsection{Comparison with different input sequences combination}
\begin{figure}[h]
  \centering
  \subfloat[]{\includegraphics[width=0.5\linewidth]{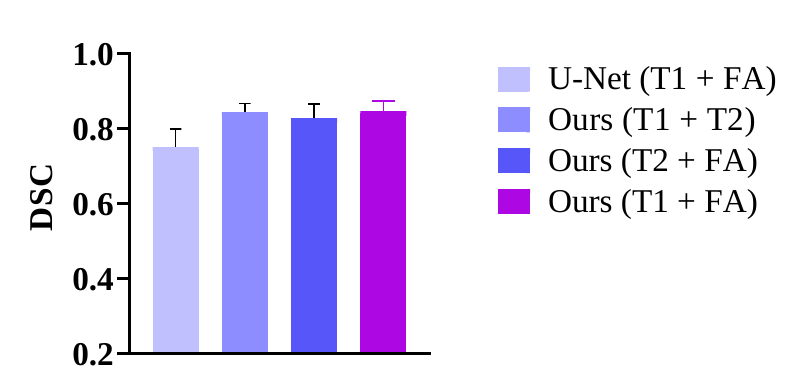}\label{fig9:a}}
  \subfloat[]{\includegraphics[width=0.5\linewidth]{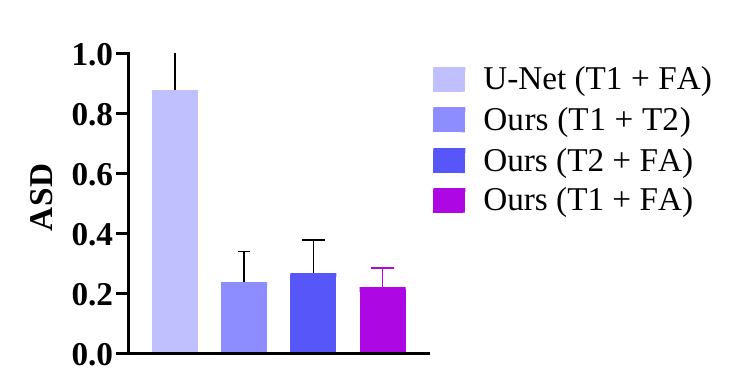}\label{fig9:b}}

   \caption{Analysis of different input sequence combinations using the HCP dataset. Subplots (a) and (b) show the DSC and ASD for our method compared to various input sequence combinations, as well as to a naïve concatenation of T1-weighted and FA images.}
  \label{fig9}
\end{figure}

In this part, we conducted a comparative analysis of our approach utilizing various combinations of input sequences, as illustrated in Fig. \ref{fig9}. Maintaining a consistent network architecture, detailed in the network structure subsection, we examined the performance across different input sequence groups. The results indicate that the combination of T1 + FA outperforms the alternatives, specifically T1 + T2 and T2 + FA, in both DSC and ASD metrics. This clear distinction underscores the superiority of our selected input combination. Additionally, we benchmarked our model against a straightforward concatenation of input sequences using the U-Net architecture, referred to as {U-Net (T1 + FA)}. The findings presented in Fig. \ref{fig9} further validate the robustness of our proposed approach, demonstrating its effectiveness in outperforming the naive concatenation strategy.

\begin{figure}[!t]
  \centering
  \subfloat[]{\includegraphics[width=0.5\linewidth]{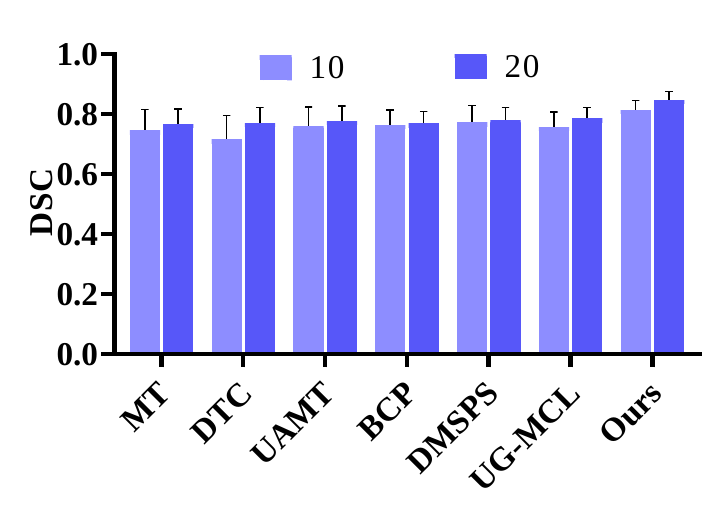}\label{fig10:a}}
  \subfloat[]{\includegraphics[width=0.5\linewidth]{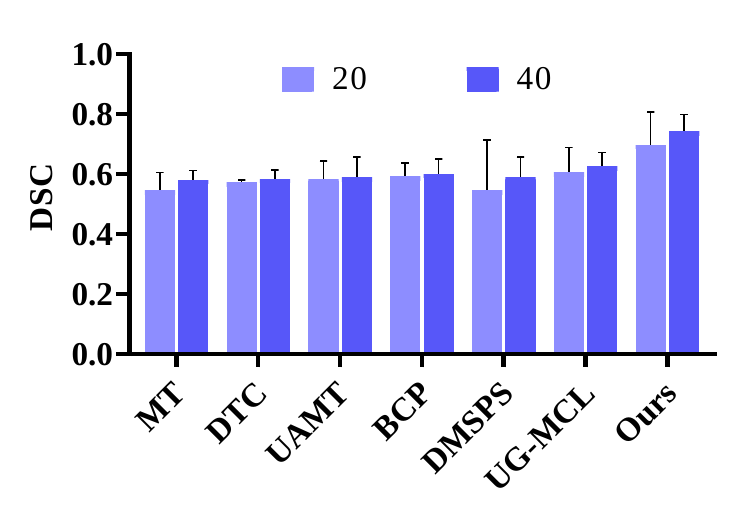}\label{fig10:b}}
   \caption{Comparison results with different labeling ratios. 10\%  and 20\% annotated HCP represent 8 and 16 labeled sample data (a), and 20\% and 40\% represent around 1 and 3 labeled MDM sample data (b).}
\label{figure10}
\end{figure}

\subsubsection{Model Complexity}
In this section, we present a comprehensive analysis of the complexity associated with various models utilized for medical image delineation. This analysis focuses on key metrics such as parameter count, Multiply-Accumulate Operations (MACs), and delineation performance, which is assessed using the DSC. Notably, all evaluated methods were trained using a semi-supervised approach. The parameter counts and MACs were computed based on an input size of $128 \times 160$ on an NVIDIA GeForce RTX 3090 GPU using the PyTorch framework and HCP dataset. As illustrated in Table \ref{tab7}, the semi-supervised (SSL) methods, which utilize the U-Net architecture mentioned earlier, have a higher parameter count of 31.04 million and a lower MAC count of 273.52 billion compared to our method. Our proposed model stands with 26.72 million parameters and 294.86 billion MACS. Despite its higher MACs compared to SSL methods, our method provides a substantial improvement of 6\% in DSC relative to the best-performing method (UG-MCL). 

These findings underline the fact that while higher parameter counts and MACs can correlate with improved performance, they also lead to increased computational demands. Therefore, the choice of model should involve careful consideration of the trade-offs between complexity and performance based on the specific requirements of clinical applications as per the conclusion of this study \cite{pang2023slim}. Our results indicate that our method achieves enhanced delineation performance even with a modest increase in MACs, making it a promising candidate for practical VP delineation tasks. 

\begin{table}[!htb]
  \centering
  \caption{Analysis of Parameters and MACS for different models with input size of $128 \times 160$ and training labeling ratio of 20\%}
  \begin{tabular}{cccccc}
    \hline
    Method & $T1$ & $FA$ & DSC  & Params (M) & MACS (G) \\
    \hline
    MT & \checkmark & \checkmark & 0.77 $\pm$ 0.05  & 31.04 & 273.52 \\
    UA-MT & \checkmark & \checkmark & 0.78 $\pm$ 0.05  & 31.04 & 273.52 \\
    DTC & \checkmark & \checkmark & 0.77 $\pm$ 0.05  & 31.04 & 273.52 \\
    UG-MCL & \checkmark & \checkmark & 0.79 $\pm$ 0.03  & 31.04 & 273.52 \\
    BCP & \checkmark & \checkmark & 0.77 $\pm$ 0.04  & 31.04 & 273.52 \\
    Ours & \checkmark & \checkmark  & \textbf{0.85 $\pm$ 0.03} & \textbf{26.72} & 294.86 \\
    \hline
  \end{tabular}
  \label{tab7}
\end{table}

\subsubsection{Delineation performance using different annotated data ratios}
To evaluate the effectiveness of our approach under different annotation ratios, we conducted a comprehensive analysis using the HCP dataset at 10\% and 20\% annotated images, and the MDM dataset at 20\% and 40\%. We meticulously plotted the delineation performance distributions in a bar chart format, incorporating standard deviations to illustrate variability for our method in comparison to six prominent semi-supervised techniques.

The findings, as depicted in Fig. \ref{figure10}, were derived from training our model with carefully optimized hyperparameters tailored for both the HCP Fig. \ref{fig10:a} and MDM Fig. \ref{fig10:b} datasets as mentioned in the training details section. Remarkably, our method consistently outperforms all semi-supervised counterparts across both datasets and at each annotation ratio, as evidenced by the higher DSC achieved. This demonstrates the robustness and superiority of our approach, emphasizing its capability to leverage limited labeled data more effectively than existing methods.  


\subsubsection{Evaluation of different features impact on the final delineation stage}
\begin{figure}[h]
\centering
\includegraphics[width=1.\columnwidth]{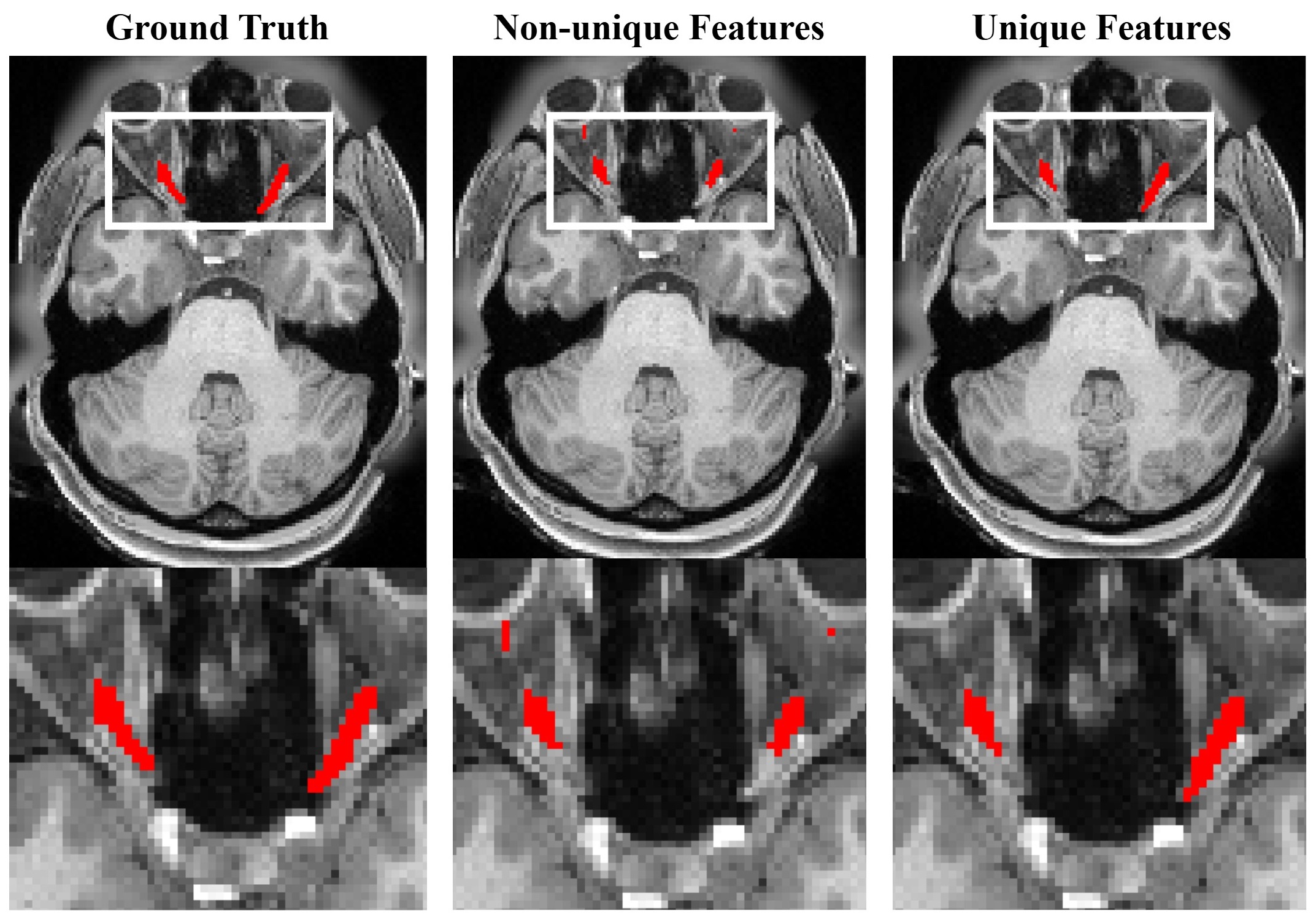} 
\caption{Visualization of different feature contributions on the HCP dataset.}
\label{figure11}
\end{figure}

In this part, maintaining a consistent network architecture, detailed in the network structure subsection, we conducted a comparative analysis of our approach utilizing unique and non-unique features on the HCP dataset, as illustrated in Fig. \ref{figure11}. The results presented in Fig. \ref{figure11} clearly demonstrates that utilizing only unique features in the final stages of the network enhances the accuracy of delineating the visual pathway structure, specifically the optic nerve, yielding a well-defined boundary. Conversely, although non-unique features improve the network's optimization through enhanced low-level feature extraction, their inclusion in the final stages adversely affects delineation performance, leading to a noticeable decline in performance. 


\subsubsection{Evaluation on the multimodality brain MRI dataset}
To further validate the method's performance on additional datasets and assess its generalizability, we conducted a thorough assessment using the multimodal brain MRI dataset with the same optimal hyper-parameters used to train the HCP and MDM datasets. Among its subjects, 10 were successfully labeled utilizing the tractography techniques outlined in this study \cite{Xie2023}. For our analysis, we strategically divided these 10 subjects into two subsets: a training set consisting of 2 labeled subjects and 6 unlabeled subjects, and a testing set consisting of 2 subjects.

\begin{figure}[h]
\centering
\includegraphics[width=0.5\linewidth]{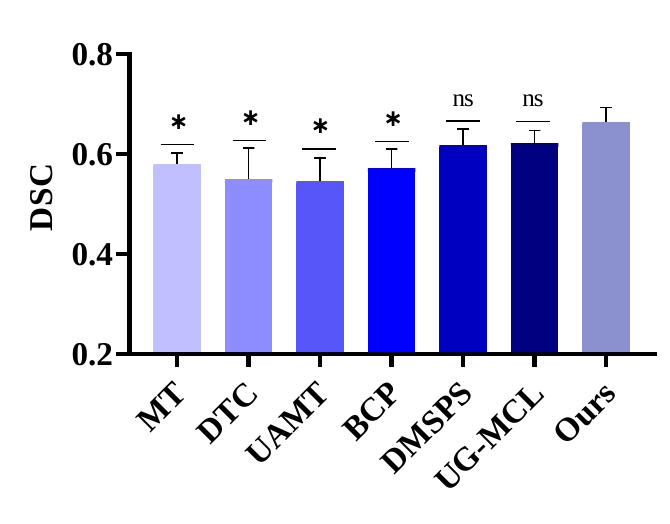} 
\caption{Statistical analysis between our method and the competing methods in terms of DSC values on the multimodality brain MRI dataset. An asterisk was used to signify $p < 0.05$, and two asterisks for $p < 0.01$. Note that since this dataset has few test samples, therefore, we run two Monte Carlo simulations before computing the paired t-test.}
\label{figure12}
\end{figure}

The experimental results, detailed in Fig. \ref{figure12}, indicate that our model significantly performs better than six (6) existing methodologies in terms of DSC metric. Notably, we achieved an impressive absolute improvement of up to 10\% in the DSC metric, reinforcing the robustness and superiority of our approach in the VP delineation task.

\section{Discussion and Conclusion} \label{discussion}

In this study, we introduced an innovative semi-supervised multi-parametric framework for the delineation of the visual pathway (VP), effectively addressing the inherent challenges of precise VP segmentation. Our approach capitalizes on the strengths of multi-parametric magnetic resonance (MR) imaging techniques while judiciously incorporating unlabeled data to improve delineation accuracy and reduce the reliance on extensive manual annotation processes.

The performance comparison with existing delineation methods on the HCP, MDM, and multimodal brain MRI datasets highlights the robustness of our approach. Our approach consistently surpasses state-of-the-art semi-supervised learning methods across all evaluation metrics, including DSC, HD95, and ASD. This highlights the superiority of our approach in achieving accurate delineation results. Furthermore, the visualization of our delineation results provides compelling evidence of the efficacy of our proposed method. We observed that our framework not only delineated the VP more accurately but also consistently identified a greater number of VP pixels across multiple datasets in comparison to the SOTA models. Particularly remarkable is our method's enhanced performance in delineating intricate structures in the optic nerve and optic tract, as demonstrated in comparisons with competitive methods, including DTC \cite{luo2021semi}, UG-MCL \cite{zhang2023uncertainty}, and DMSPS \cite{han2024dmsps}. Moreover, our method performs better than fully supervised methods in scenarios where the available annotated samples are very limited. These observations solidify our argument regarding the robustness of our method, showcasing its capacity to yield consistent and discriminative results across diverse datasets using limited annotated samples.

Nevertheless, it is imperative to acknowledge certain limitations within our study. Notably, our current focus is predominantly on the delineation of the VP. In the future, we may explore integrating additional information and advanced analysis techniques to provide a more comprehensive understanding of the visual pathway, such as quantifying structural changes or functional analysis. This has the potential to improve diagnostic protocols and treatment planning for patients with visual disorders \cite{he2023unified}. Additionally, by assessing both the structural and functional aspects of the visual pathway, clinicians are expected to be better equipped to tailor interventions and monitor therapeutic effects over time.

To further establish the performance and generalizability of our approach, we advocate for larger-scale studies encompassing multiple institutions and diverse patient populations. Such efforts would provide robust validation of our method's efficacy. Collaboration with specialists in ophthalmology and neurology could also yield invaluable insights, facilitating the transition of our innovative method into routine clinical practice.

\bibliographystyle{IEEEtran}

\end{document}